\documentclass[12pt]{article}
\usepackage{inputenc}
\usepackage{booktabs}
\usepackage{multirow}
\usepackage{float}
\usepackage{cite}
\usepackage{amsmath,amssymb,amsfonts}
\usepackage{algorithmic}
\usepackage{textcomp}
\usepackage{indentfirst}
\usepackage{fancyhdr}
\usepackage{url}
\usepackage{subfigure}
\usepackage{comment}
\usepackage{caption}
\usepackage{tabularx}
\usepackage{geometry}
\usepackage{color}

\newcommand{\Team}{------}
\usepackage{amsthm}
\usepackage{newtxmath} % must come after amsXXX

\usepackage{graphicx}
\begin{document}
\graphicspath{{.}}  % Place your graphic files in the same directory as your main document
\DeclareGraphicsExtensions{.pdf, .jpg, .tif, .png}
\thispagestyle{empty}
\vspace*{-16ex}
\centerline{\begin{tabular}{*3{c}}
	\parbox[t]{0.3\linewidth}{\begin{center}\textbf{2021\\ MCM/ICM\\ Summary Sheet}\end{center}}
\end{tabular}}
%%%%%%%%%%% Begin Summary %%%%%%%%%%%
% Enter your summary here replacing the (red) text
% Replace the text from here ...

~\\
\centerline{\large\textbf{Asian Giant Hornet Control based on Image Processing and Biological Dispersal}}
~\\
Changjie Lu, Shen Zheng, Hailu Qiu 
\\College of Science and Technology, Wenzhou-Kean University
\vspace{0.4cm}
\\
\centerline{\textbf{Summary}}
The Asian giant hornet (AGH) appeared in Washington State appears to have a potential danger of bioinvasion. Washington State has collected public photos and videos of detected insects for verification and further investigation. In this paper, we analyze AGH using data analysis, statistics, discrete mathematics, and deep learning techniques to process the data to control AGH spreading.
\par
First, we visualize the geographical distribution of insects in Washington State. Then we investigate insect populations to varying months of the year and different days of a month. Third, we employ \textbf{wavelet analysis} to examine the periodic spread of AGH. Fourth, we apply  \textbf{ordinary differential equations} to examine AGH numbers at the different natural growth rate and reaction speed and output the potential propagation coefficient. Next, we leverage \textbf{cellular automaton} combined with the potential propagation coefficient to simulate the geographical spread under changing potential propagation. To update the model, we use \textbf{delayed differential equations} to simulate human intervention. We use the time difference between detection time and submission time to determine the unit of time to delay time. After that, we construct a lightweight CNN called \textbf{SqueezeNet} and assess its classification performance. We then relate several \textbf{non-reference image quality metrics}, including NIQE, image gradient, entropy, contrast, and TOPSIS to judge the cause of misclassification. Furthermore, we build a \textbf{Random Forest} classifier to identify positive and negative samples based on image qualities only. We also display the feature importance and conduct an error analysis. Besides, we present sensitivity analysis to verify the robustness of our models. Finally, we show the strengths and weaknesses of our model and derives the conclusions.

We find meaningful conclusions from our analysis. 
(1) Most Asian Giant Hornet (AGH) is reported at West Washington from May to December. However, only 14 AGH proves to exist.
(2) Local insects have a monthly active cycle of about 7.71 days. AGH breed at a rapid rate every Spring. Meanwhile, the geographical spread of AGH heavily depends on the potential propagation coefficient.
(3) Earlier artificial intervention would efficiently control the AGH population and protect other insects.
(4) Light-weight convolutional neural network is efficient and robust for classifying AGH.
(5) High image Gradient Pictures tend to contain AGH, while high image Contrast and background interference will misguide classification.

We finally provide recommendations based on our analysis result. 
(1) AGH has a tiny population and is nearly eradicated in Washington State. Therefore is no need for massive panic. 
(2) AGH has a predicted spread period of approximately one and a half months. When their growth rate is high, we should take earlier actions and eliminate them before expansion. 
(3) We should prioritize image gradient over image contrast and ignore background noise for subjective classification. 
(4) We recommend a lightweight CNN for accurate classification. 

~

% to here
%%%%%%%%%%% End Summary %%%%%%%%%%%

%%%%%%%%%%%%%%%%%%%%%%%%%%%%%%
\clearpage
\pagestyle{fancy}
% Uncomment the next line to generate a Table of Contents
%\tableofcontents 
\newpage
\setcounter{page}{1}
\rhead{Page \thepage\ }
%%%%%%%%%%%%%%%%%%%%%%%%%%%%%%
\newpage
\tableofcontents
\newpage
\setcounter{page}{2}

\section{Introduction}
\subsection{Background}

Asian Giant Hornet (Vespa mandarinia), a hornet species native to eastern Asia's temperate and tropical areas, is a predator to pollinating insects, the world's most enormous hornets.
\par
On Vancouver Island in British Columbia, Canada, Sep 2019, an AGH colony was detected and immediately destroyed. In May 2020, an AGH individual was discovered in Washington state, which means AGH survived a winter in northern America.
\par
In the 21st century, biological invasions have become one of the most significant problems to biodiversity, crop production, and ecological balance. As one kind of social Hymenoptera, AGH perfectly fits in a spreading pattern by human transportation and hidden skills from standard border inspection. They hide in transported goods and can quickly establish a new colony with only one fertilized queen. Furthermore, high fertility, comprehensive diet and habitat, great defense, and competitive abilities also make them dominant in the area\cite{beggs2011ecological}. A species under the same genus with AGH, which is called V. velutina, invaded throughout the European continent with a rapid speed, and their control costs can be very high\cite{alaniz2021giants}. A laboratory researched AGH spreading pattern by building habitat suitability models and dispersal simulations, concluded that AGH has a high possibility of spreading across western North America with no human intervention\cite{zhu2020assessing}.
\par
To investigate the current situation of AGH and find the government of Washington State offered a solution defending its bioinvasion, hotlines, and a website for people to report the hornets' emerges in different areas. The public reports need to be processed through multiple analyses for further study and giving solutions.

\subsection{Related Works}

Convolutional neural network (CNN) has been widely applied to image classification tasks because of their excellent accuracy, capability, and flexibility. Krizhevsky \cite{krizhevsky2012imagenet} introduces AlexNet, which uses max-pooling layers and drop-out techniques. Simonyan \cite{simonyan2014very}, with others in VGG (Visual Geometry Group), designed very deep convolutional networks for large-scale image recognition, which grows deeper layers with a small kernel size of 3 * 3. Lightweight image classification models have been designed to fit mobile devices that have limited memory capacity. MobileNet \cite{howard2017mobilenets} uses depth-wise separable convolutions to a trade-off between speed and accuracy efficiently. SqueezeNet \cite{iandola2016squeezenet} adopts small filters, late downsampling, and fire block (squeeze and expand operations) for flexible and efficient model compression.

\subsection{Solutions}
In this paper, we propose a novel approach for analyzing, predicting, and giving recommendations towards Asia Hornet.  Our analysis uses different discrete mathematics, deep learning, and data visualization tools.
First, we import and preprocess the data, including mapping images, videos to text, and numbers.
Second, we use dispersal models. Our dispersal model mainly includes different equations, cell automation (CA), and wavelet analysis. 
Third, we bring a CNN for classifying AGH. We leverage statistical evaluation metrics and activation maps to verify our model performance.
Fourth, we explore the cause for misclassification using a random forest \cite{breiman2001random} classifier with several non-reference image quality metrics.\par
\begin{figure}
    \centering
    \includegraphics[width =15cm]{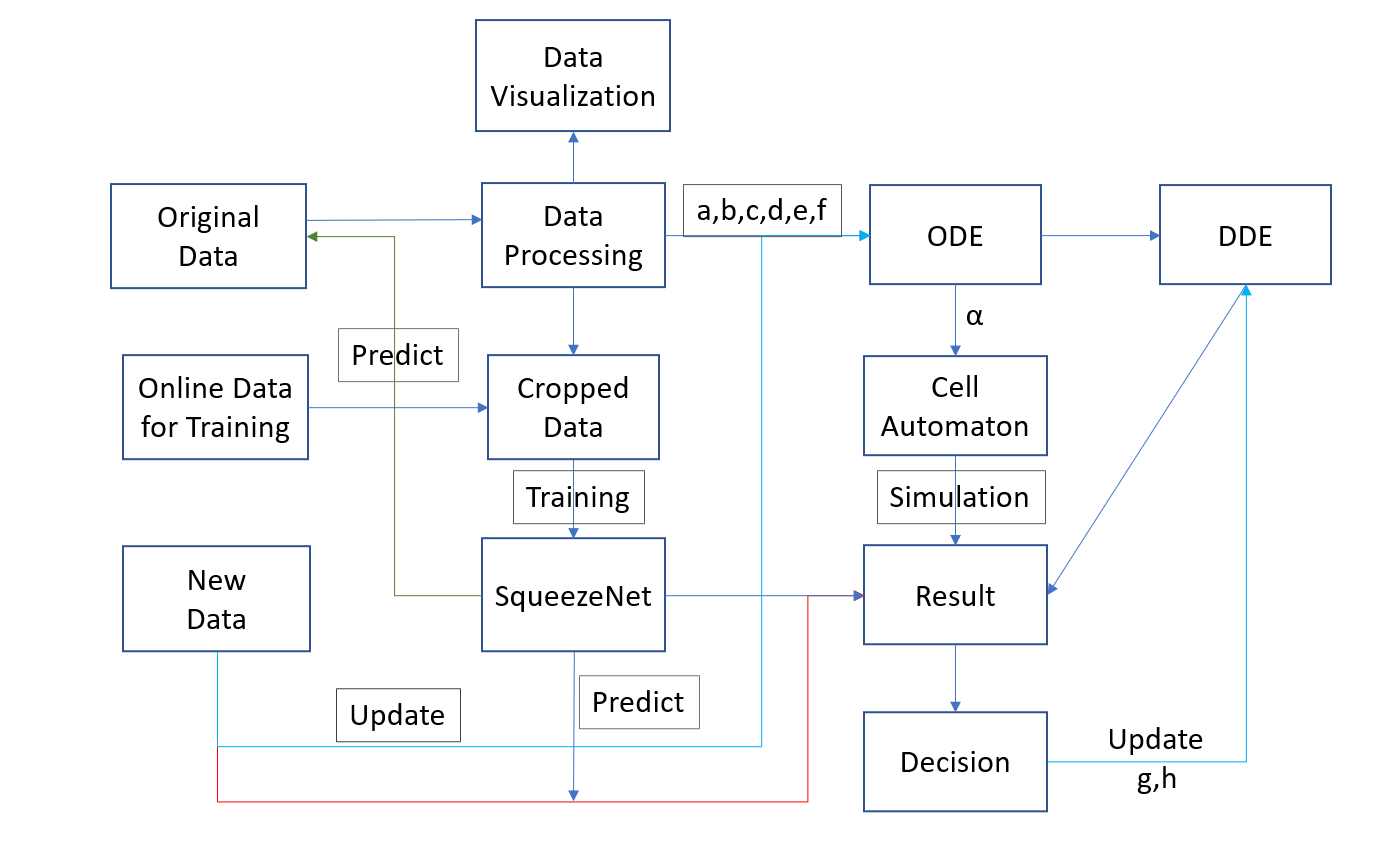}
    \caption{\textbf{Paper Workflow}}
    \label{fig:my1}
\end{figure}
The rest of the paper is organized as below. Section 2 introduces basic notations and assumptions. Data acquisition and data visualization are conducted in Section 3. Section 4 and section 5 construct the expansion model and the classification model, respectively. Section 6 perform sensitivity analysis and Section 7 analyzes the strengths and weaknesses. Conclusion and recommendations will be addressed in the Memorandum.

\section{Assumptions and Notations}
\begin{table}[htp]
\centering{
\begin{tabular}{llll}
\toprule[1.5pt]
 & Abbreviation & Full form                                            &  \\ \hline
 & AGH      & Asian Giant Hornet                                   &  \\
 & CNN      & Convolutional Neural Network                                                         &  \\
 & CA      & Cellular Automaton                                           &  \\
 & RF      & Random Forest
                            &  \\
 & ODE      & Ordinary Differential Equation
                            &  \\
 & DDE      & Delay Differential Equation
                            &  \\
 & TOPSIS      & Technique for Order of Preference by Similarity to Ideal Solution          &  \\
 & NIQE      & Naturalness Image Quality Evaluator
                            &  \\
 & ReLU      & Rectified Linear Unit
                            &  \\
 & Conv2d      & 2D Convolution Layer
                            &  \\
 & OOB      & Out Of Bag
                            &  \\

\bottomrule[1.5pt]

\end{tabular}
}
\caption{\textbf{Abbreviation of Relevant Terms}}
\end{table}

\begin{table}[h]
\centering{

\begin{tabular}{llll}
\toprule[1.5pt]
 & Symbol & Explanation                                                                                                                                     &  \\ \hline
 & A(t)      & Amount of AGH at any given time                                                        &  \\
 & B(t)      & Amount of local species at any given time                                               &  \\
 & a      & Natural growth rate of AGH                                            &  \\
 & b      & Internal competition coefficient of AGH
                            &  \\
 & c      & Aggression coefficient of AGH to local species
                        &  \\ 
 & d      & Natural growth rate of local species
                            &  \\
 & e      & Internal competition coefficient of local species
                            &  \\
 & f      & Resistance coefficient of local species to AGH
                        &  \\ 
 & g      &Effects of manual intervention on AGH
 &\\
 & h     &Effects of manual intervention on local species
 &\\
\bottomrule[1.5pt]
\end{tabular}
}
\caption{\textbf{Symbols and Explanations}}
\end{table}

\begin{itemize}
    \item We assume that AGH only competes with local species and receive mininum impact from altitude, weather, and human activities.
    \item We assume that the map of Washington State can divides into 128*128 squares.
    \item We assume that the spread pattern of AGH in Washington State is relatively fixed.
    \item We assume that the proportional relationship between AGH and other dwelling species is similar to the number of reports submitted within a specific range of latitude and longitude.
\end{itemize}

\section{Data Preprocessing}
\subsection{Data Acquisition}
We have several data files for analysis. The first is a pdf file named Vespamandarinia, which we have used in the introduction section. The second is a pair of a spreadsheet. We use GlobalID to combine two spreadsheets. We first convert all images into jpg format. We use the strategy for word and pdf file that contains only images. We then take a screenshot of the video and convert it to jpg images. We take the screenshot at the first explicit appearance of insects.

\subsection{Data Visualization}

\begin{figure}[H]
    \centering
    \includegraphics[width = 14cm]{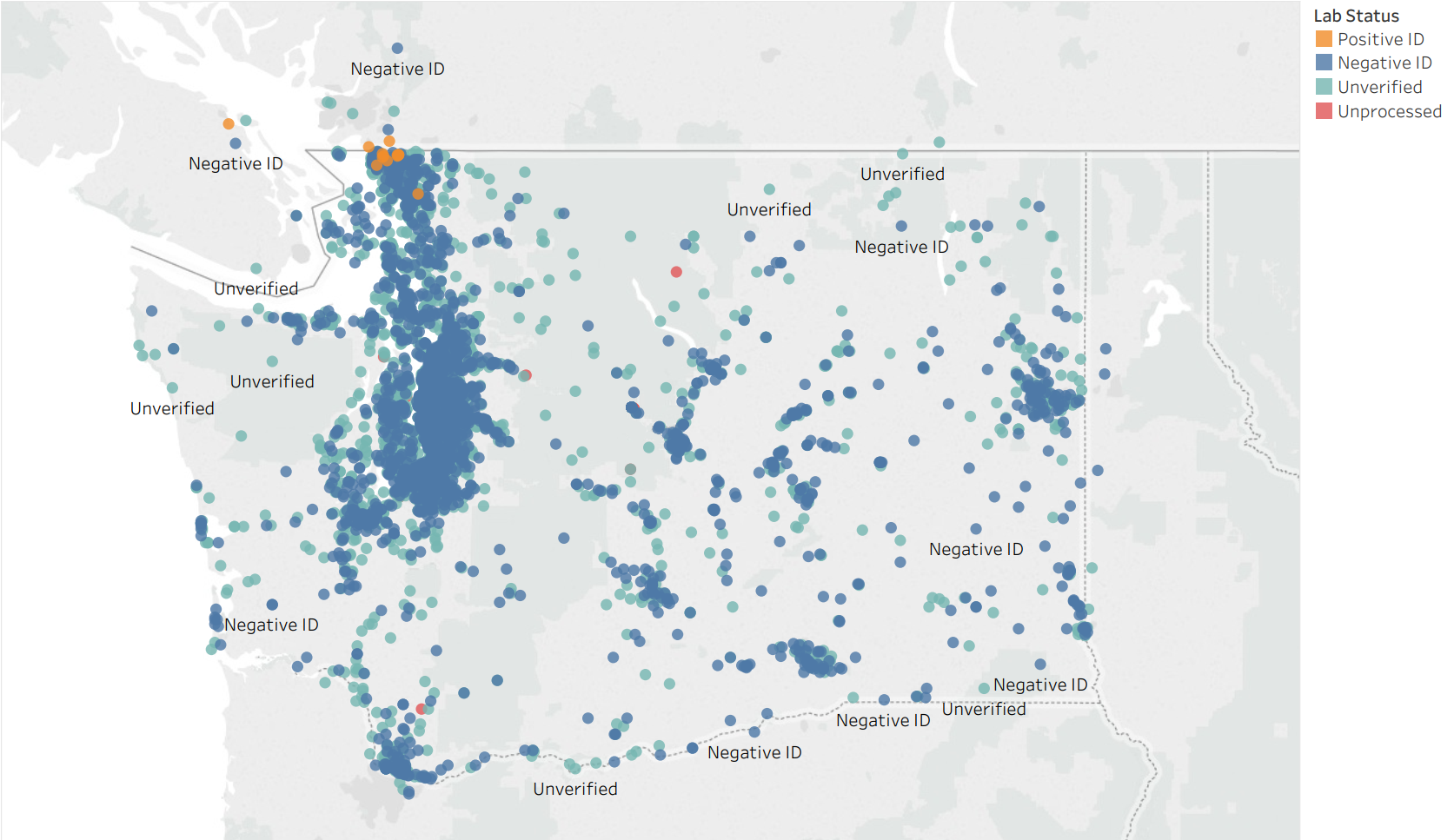}
    \caption{\textbf{Geographical Visualization}}
    \label{fig:my2}
\end{figure}

We perform classification using lab comments and lab status. We count each insect species' number under the same lab status in different years by detection date, from January to December, integrating them into one spreadsheet. The Geographical Visualization result in Fig.1 shows most reported cases appear in west Washington. Besides, all positive ID appears in northwest Washington.

% Change of Insect Population over Time
\begin{figure}[H]
    \centering

    \includegraphics[width=16cm]{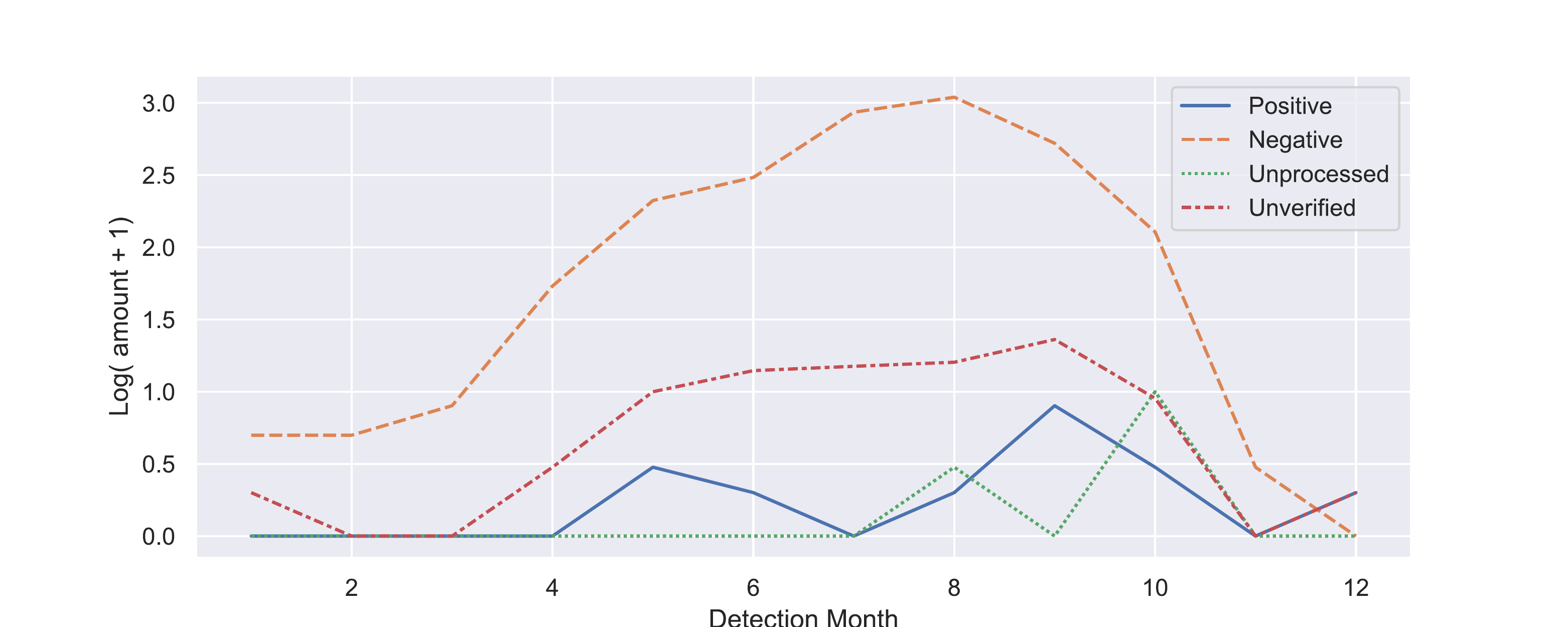}
    \caption{\textbf{Change of Insect population over Time.} X-axis: the month where a insect is observed. Y-axis: Corrected log value of insect amount in a specific month. The category positive, negative, unprocessed, and unverified from lab status data means AGH species insects, non-AGH species insects, unprocessed insects, and unverified insects respectively.}
    \label{fig:3_2_Total.pdf}
\end{figure}

We have several findings from Fig.3. First, each category begins a rapid growth in May and peaked in around September. 
Second, the negative category has the highest numbers. 
Third, the positive and unprocessed category has the least numbers with the similar trend, where as the unverified category has a similar trend with the negative category. 

In Fig.4(b), we have categories including symphyta (sawfly), wasp, Sphex ichneumoneus (great golden sand digger), Bombus (bumblebee) and Apis (honey bee), Dolichovespula maculata (bald-head hornet) and Vespa crabro (European hornet), and other species. 

% Change of AGH population over time
\begin{figure}[H]
    \centering
    \subfigure[]{
        \includegraphics[width=14cm]{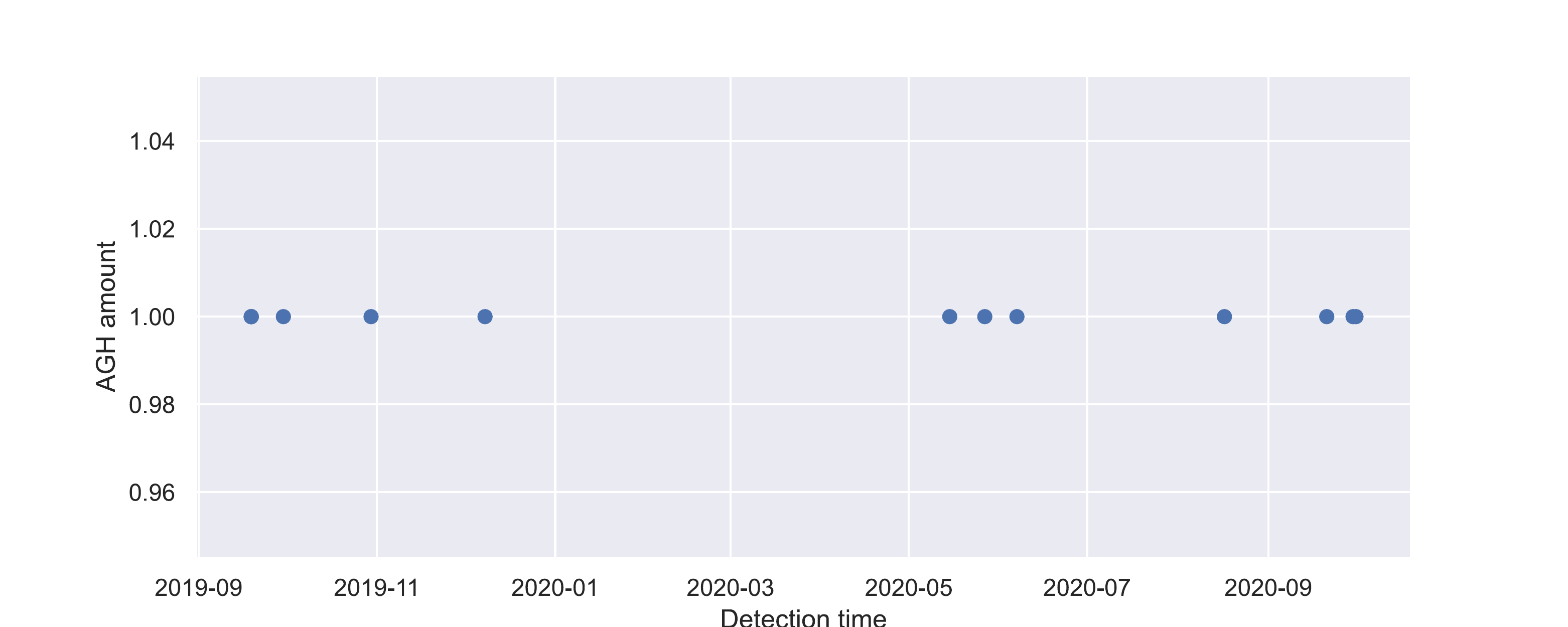}
        }
    \subfigure[]{
        \includegraphics[width=14cm]{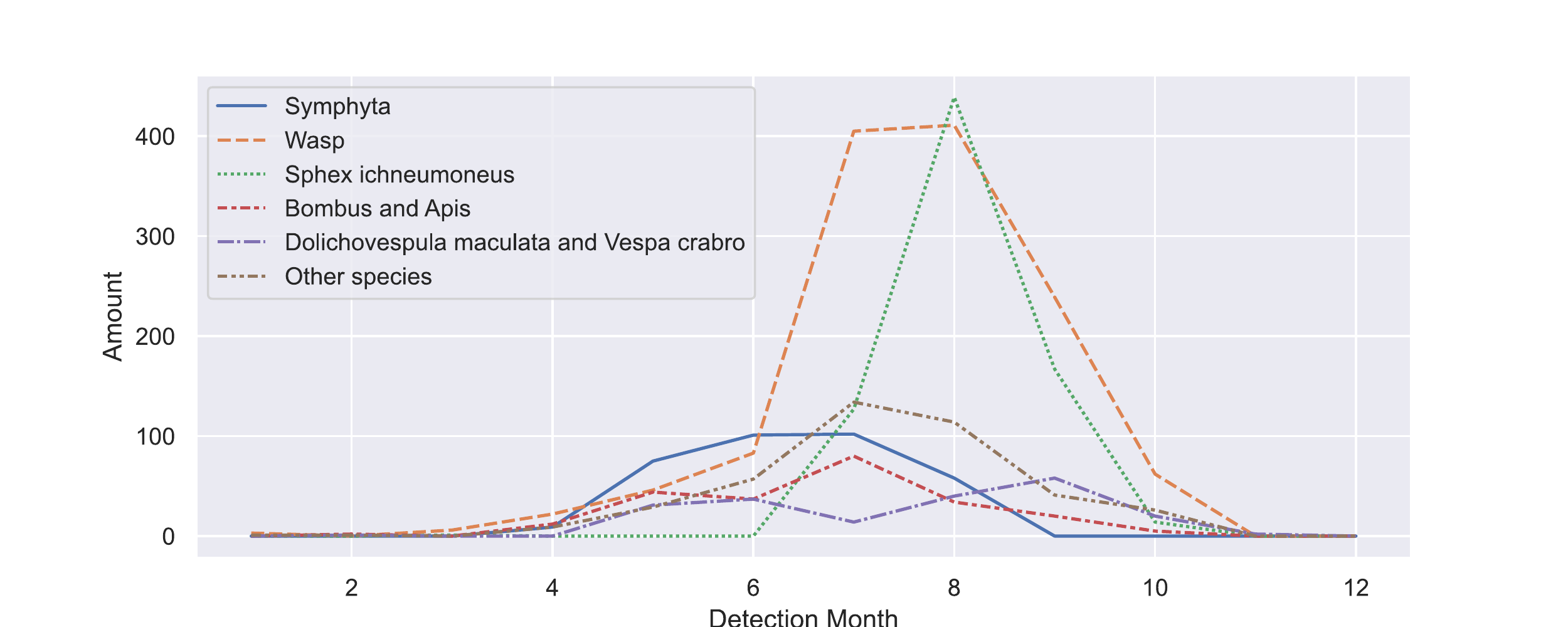}
        }
    \caption{(a): \textbf{Change of AGH population over time.} X-axis: the year and month when a AGH is observed. Y-axis: Numbers of AGH at a specific time. (b): \textbf{Change of Non-AGH population over time}. X-axis: the month where a insect is observed. Y-axis: the numbers of  reported insects that is not AGH. We roughly classified insects into 5 categories and others.}
    \label{fig:3_2_Positive.pdf}
\end{figure}
We have several findings from Fig.4. 
First, there are 14 AGHs detected on specific dates. 
Second, no AGHs is detected during 2020-01 to 2020-05.
Third, only one AGD is reported in one detection.
Besides, all non-AGH insects have a rapid increase in June and a signficant decline in September. 
Last, Sphex ichneumoneus and wasp have peak values of around 400 during July and August, whereas others category of insects does not.

\begin{figure}[H]
    \centering
    \includegraphics[width = 14cm]{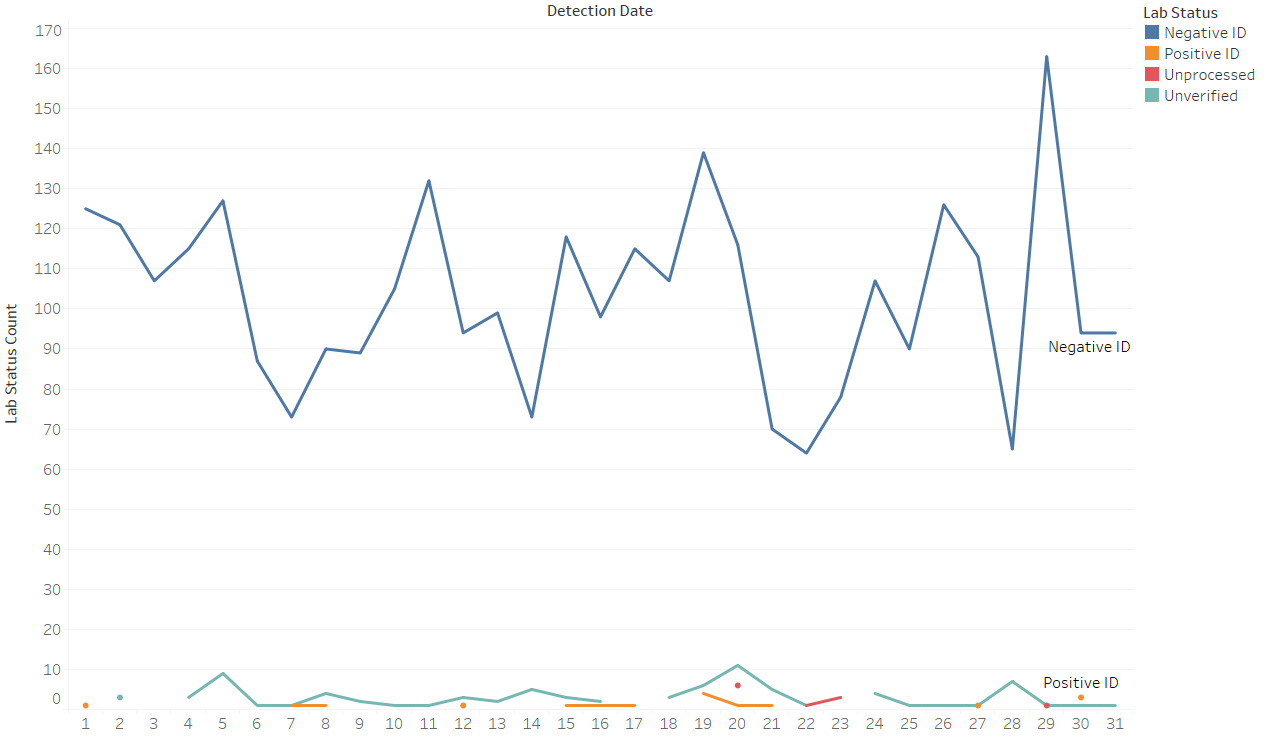}
    \caption{\textbf{Total Insect Numbers Detected by Day in a Month.} X-axis: day in a month, from 1 to 31. Y-axis: lab status comment for indicating insect categories. Legend: different lab status}
\end{figure}
    
\begin{figure}[H]
\centering
    \includegraphics[width = 14cm]{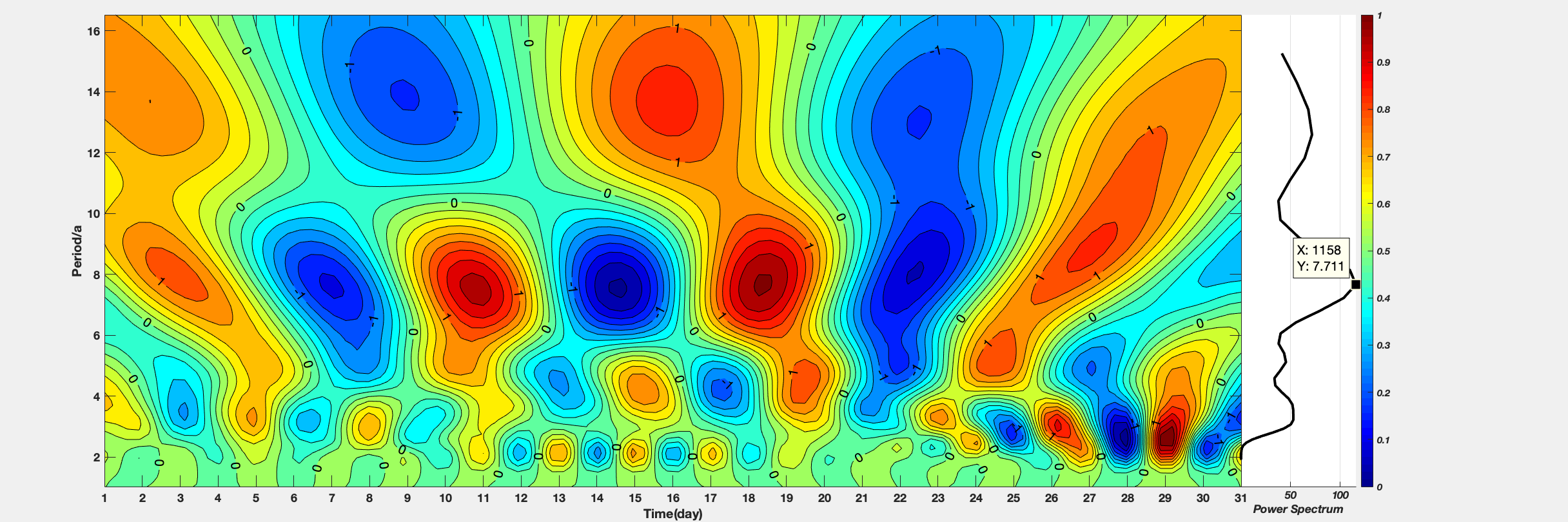}
    \caption{\textbf{Insect Periodic Activity with Wavelet Analysis.} Left X-axis: time in days since simulation. Left Y-axis: cycle time period. Right X-axis: power spectrum frequency. Right Y-axis: intensity}
\end{figure}

We introduce wavelet analysis \cite{torrence1998practical} in Fig.6 to investigate if the insects have monthly activity patterns. Wavelet analysis figure show that inserts have a 7.71-day period, which explains Washington State insects' monthly activity as 7 days. 
%More can be sampled at the peak of activity of insects to improve screening efficiency.

\section{Expansion Model}

\subsection{Differential Equations}
\subsubsection{Ordinary Differential Equations}

We use systems of ordinary differential equations (ODE) to simulate how insect populations evolve. Our different equations seek equilibrium points given initial values and constraints. In Fig.7, we describe AGH as competing species to local species.

We set up a differential equation to study the population change of AGH in Washington. The system of differential equations established is as follows:
\begin{equation}
    \left\{\begin{array}{l}
\frac{d A(t)}{d t}=a A(t)-b A^{2}(t)-c A(t) B(t) \\
\\
\frac{d B(t)}{d t}=d B(t)-e B^{2}(t)-f A(t) B(t)
\end{array}\right.
\end{equation}
% Solution of ODE 
\begin{figure}[H]
    \centering
    \subfigure[a=0.5]{
    \includegraphics[width=7cm]{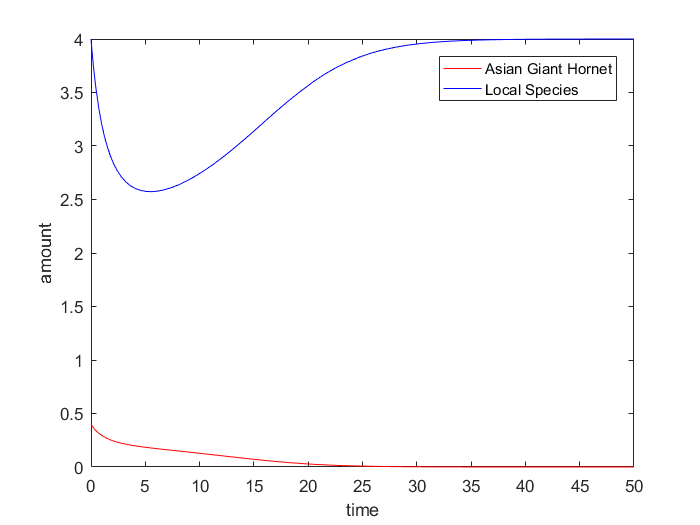}
    }
   \quad
   \subfigure[a=0.55]{
    \includegraphics[width=7cm]{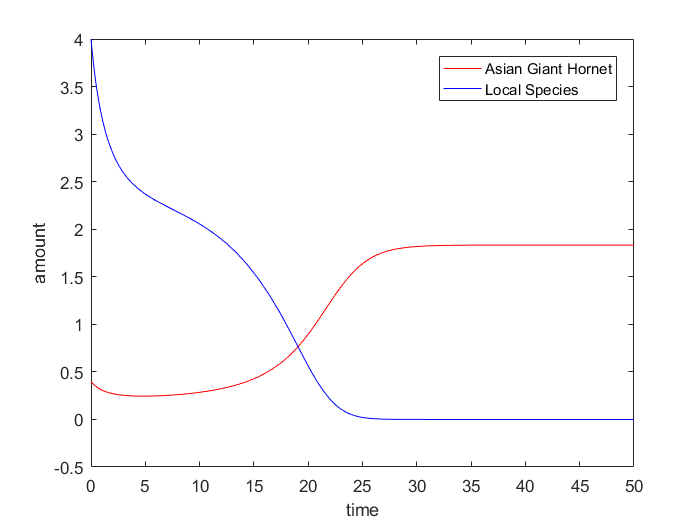}
    }
    \caption{\textbf{The solution of our ODE systems.} (a): Solution when Natural Growth Rate of Asian Giant Hornet (AGH) is 0.5. (b): Solution when growth rate is 0.5. For both conditions, we set b = 0.3, c = 0.2, d = 0.4, e = 0.1, and f = 0.8}
\end{figure}
The number of AGH found in northwest Washington is roughly 1:10 for the number of Local species. Therefore, we set A = 0.4, B = 4 as a starting point. The result of our ODE appears in Fig.7. When the natural growth rate of AGH is 0.5, the amount of AGH decreases steadily from 0.5 to 0, whereas the amount of local species decreases sharply to 2.5 before slowly increasing to 4.0. 
When the natural growth rate of AGH is 0.55, the amount of AGH increases to 2, whereas the number of local species decreases to 0. When time is greater than 25, they arrive at equilibrium.

\subsubsection{Delay Differential Equation}
We use systems of delay differential equations (DDE) to analyze the problem further. Unlike ODE, DDE has the derivative of the unknown function in terms of previous time points. The formula of our delayed differential equation is below. Note the last terms at each line comes from an early reaction. Others are the same as the ODE.\par
We calculated the average difference between image submission time and detection time. The parameters g and h represent the influence of the manual intervention on the population. (g = 0.05, h = 0.001)
Delays in manual intervention can lead to dramatic changes in population evolution.

\begin{equation}
    \left\{\begin{array}{l}
\frac{d A(t)}{d t}=a A(t)-b A^{2}(t)-c A(t) B(t)-g A(t-10) \\
\\
\frac{d B(t)}{d t}=d B(t)-e B^{2}(t)-f A(t) B(t)-h B(t-10)
\end{array}\right.
\end{equation}

\begin{figure}[H]
    \centering
    \includegraphics[width = 8cm]{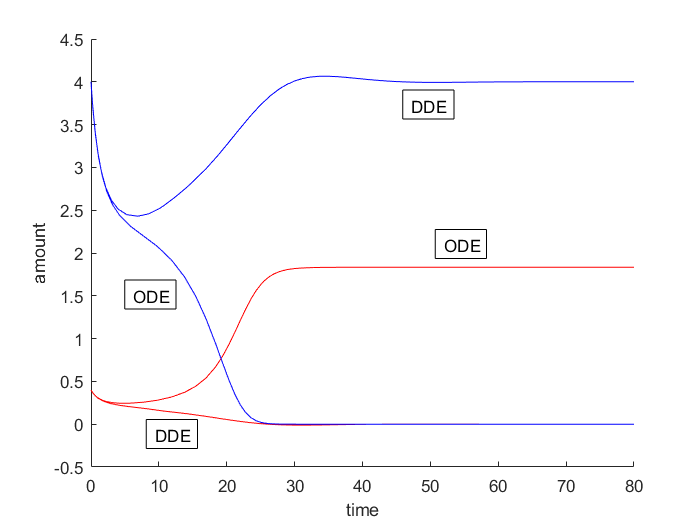}
    \caption{\textbf{Solution Comparison between ODE and DDE.} We set the natural growth rate of AGH as 0.55. We use red for AGH and blur for other species.}
    \label{fig:my3}
\end{figure}

Fig.8 shows the result from earlier and on-time actions towards AGH. In the ODE solution, there is quick increase of AGH and a sharp decrease of other species. In the DDE solution, there is quick increase of other species and steadily decline of AGH. Our result demonstrate earlier action help control AGH numbers.

%This means that if the public does not upload the images in time or the government is too slow to approve them, the good time to control the AGH may be missed.

\subsection{Cell Automaton}

Cellular Automaton (CA) is a discrete computational model in automation theory. It consists of finite-dimensional cells with states (e.g., True/False). A new generation is created according to specific algorithms to iterative update the neighboring cells. 

From the the ODE, We get values of A(t) and B(t) with respect to time. Therefore, we define the potential propagation coefficients:
\begin{equation}
    \alpha(t)=\frac{A(t)}{A(t)+B(t)}
\end{equation}
\par
We apply CA to simulate the spread of AGH in Washington and simplify the map into a square, dividing into 128*128 cells. In the upperleft 15*15 grid, we use red to represent AGH existence and black to represent native species. Since most AGH cases is reported in northwest Washington, we set initial probability ratio as 4:1. The probability of each grid occupied by AGH is (if P < 1):

\begin{equation}
    P(\text{cells }=1)=\frac{\operatorname{sum}}{2} \cdot \alpha\left(\frac{t}{10}\right)
\end{equation}

Where sum(x,y) represents AGH's amount around the grid with coordinate position (x,y), and t is the time unit of CA.
After every 50 units of time, there will be 10 units where AGH cannot reproduce and has a death probability of 0.3: 
\begin{equation}
    \text { P(cells }=0)=0.3
\end{equation}

% Solution of Cell Automation
\begin{figure}[H]
\centering
\subfigure[t=1]{
\includegraphics[width=2.8cm]{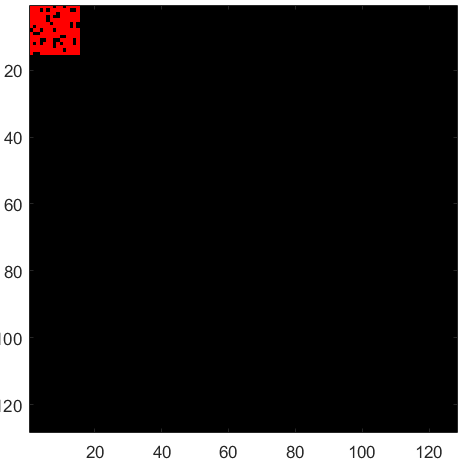}
}
\quad
\subfigure[t=11]{
\includegraphics[width=2.8cm]{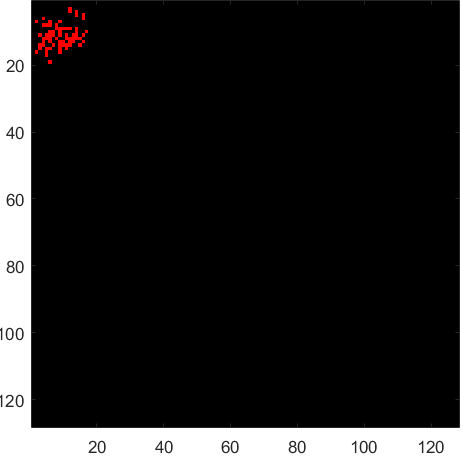}
}
\quad
\subfigure[t=21]{
\includegraphics[width=2.8cm]{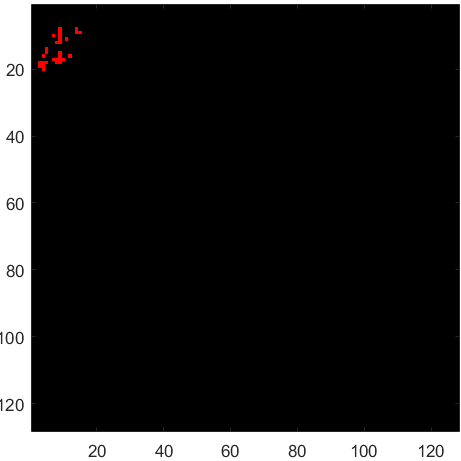}
}
\quad
\subfigure[t=50]{
\includegraphics[width=2.8cm]{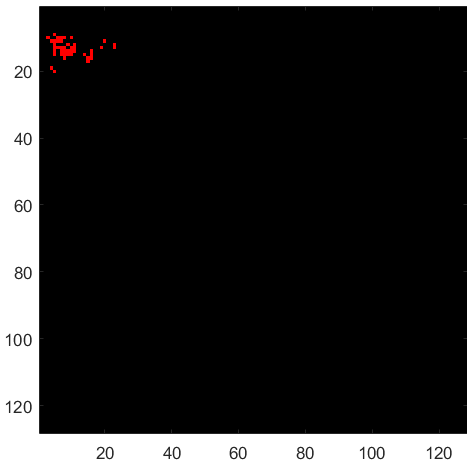}
}
\quad
\subfigure[t=64]{
\includegraphics[width=2.8cm]{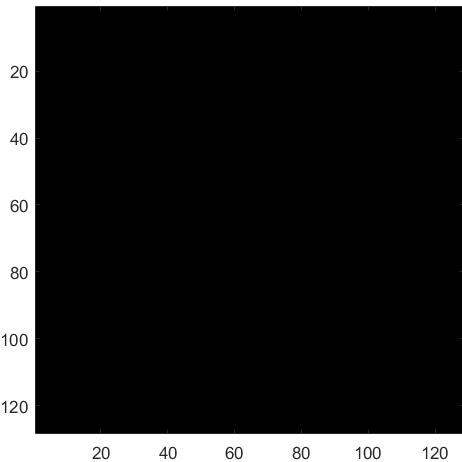}
}
\caption{\textbf{Cellular Automaton Using the First Potential Propagation Coefficient}}
\end{figure}

\begin{figure}[H]
\centering
\subfigure[t=1]{
\includegraphics[width=3cm]{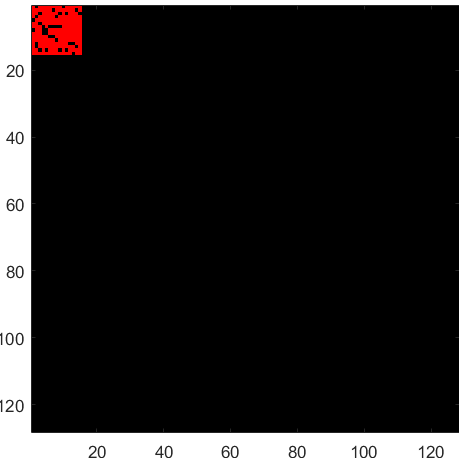}
}
\quad
\subfigure[t=12]{
\includegraphics[width=3cm]{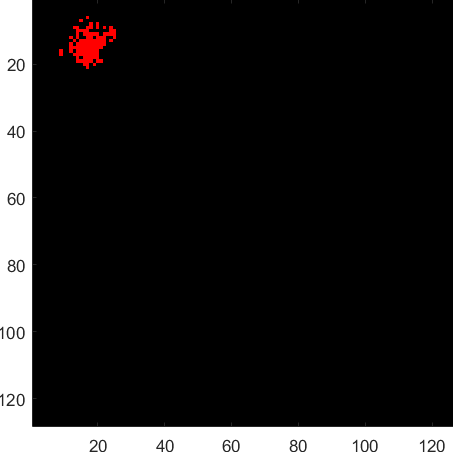}
}
\quad
\subfigure[t=67]{
\includegraphics[width=3cm]{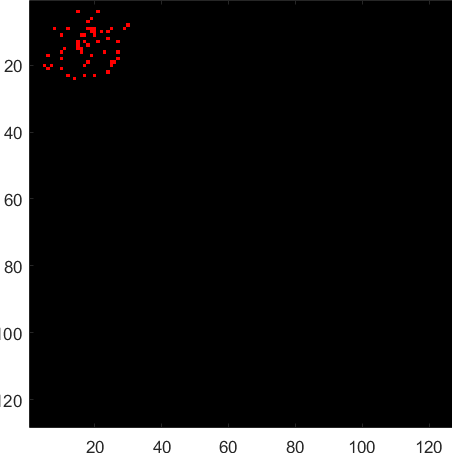}
}
\par
\quad
\subfigure[t=85]{
\includegraphics[width=3cm]{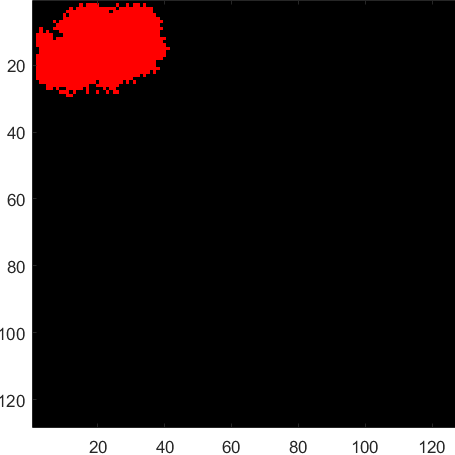}
}
\quad
\subfigure[t=153]{
\includegraphics[width=3cm]{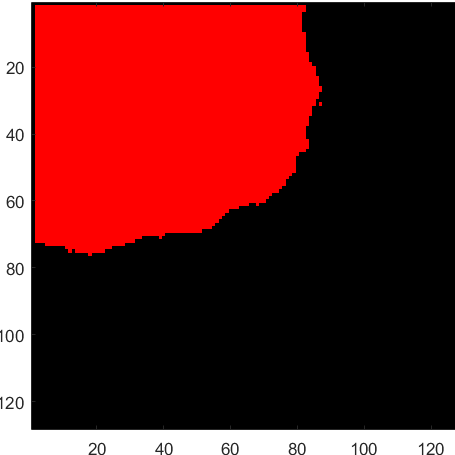}
}
\quad
\subfigure[t=163]{
\includegraphics[width=3cm]{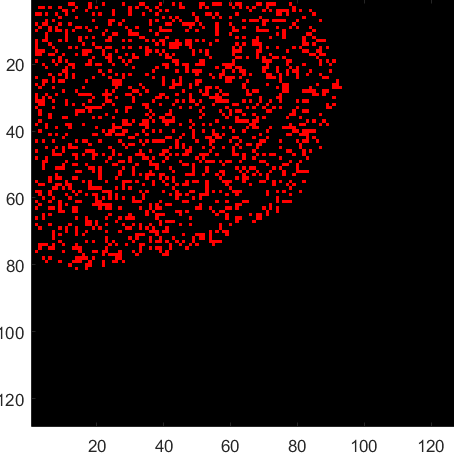}
}
\quad
\subfigure[t=241]{
\includegraphics[width=3cm]{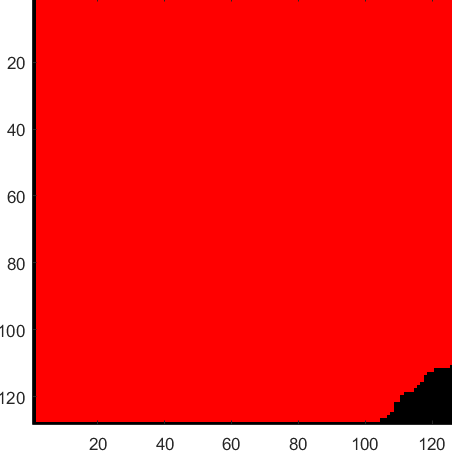}
}
\caption{\textbf{Cellular Automaton Using the Second Potential Propagation}}
\end{figure}

In Fig.9, we visualize AGH's evolution based on the first differential equation, from t = 1 to 64, AGH gradually decrease and finally disappear. 

In Fig.10, we visualize AGH's evolution based on the first differential equation, from t = 1 to 64, AGH gradually decrease and finally disappear. , we visualize AGH's evolution based on the second differential equation. From t = 1 to 67, AGH has small and fluctuating numbers. Starting from t = 85, the AGH population grow explosively and finally occupy Washington.

\subsection{Wavelet Analysis}
% Wavelet Transform Analysis Result
Wavelet transform is a decomposition of a continuous signal into different scale components. Wavelet analysis is better than Fourier analysis in that it can capture the time information of frequency change. Therefore, wavelet analysis is suitable for complicated movements such as the spread of AGH.
Our Morlet wavelet is a continuous plane wave modulated by Gauss function:
\begin{equation}
    \Psi(t)=e^{i c l}\left(\mathrm{e}^{-\frac{t^{2}}{2}}-\sqrt{2} \mathrm{e}^{-\frac{c^{2}}{4}} \mathrm{e}^{-l^{2}}\right)
\end{equation}
Its discrete mathematical expression is:
\begin{equation}
    W_{f}(a, b)=|a|^{-\frac{1}{2}} \sum_{i=1}^{N} f(i \delta t) \psi^{\star}\left(\frac{i \delta t-b}{a}\right)
\end{equation}
where * represents complex conjugate;
a is scale factor (related to period and frequency);
b is the translation factor (time position);
I is time and position label of data series;
f(t) is the variable time series;
W is the wavelet coefficient;
$\delta t$ is the time interval of the variable series.
Since the wavelet is in complex from, the coefficients after the wavelet transform are also complex.

\begin{figure}[H]
    \centering
    \includegraphics[width = 10cm]{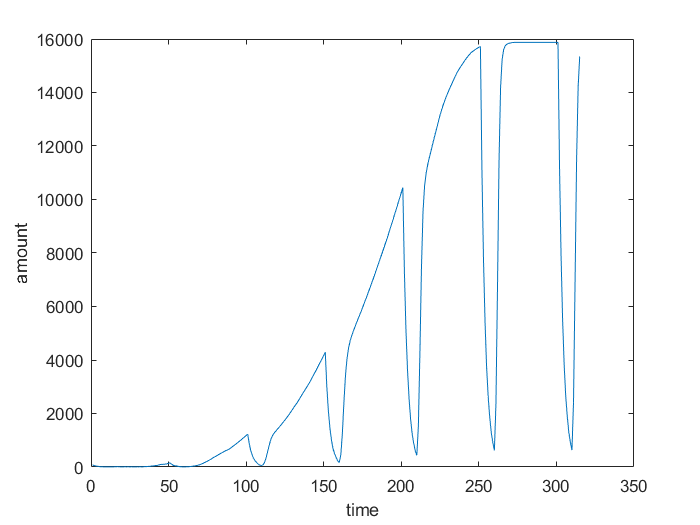}
    \caption{\textbf{Red Grid Numbers in Cellular Automaton over Time.} X-axis: Time since starting simulation. Y-axis: Red Grid amount. We select the condition of the second differential equation for plotting. More Grid counts represents more AGH occupancy.}
\end{figure}

Fig.11 shows a general increasing trend for AGH numbers. However, there is a sharp decrease for after every 50 time units. Therefore, eliminating the Asian bumblebee in winter is a better option.

\begin{figure}[H]
    \centering
    \includegraphics[width = 14cm]{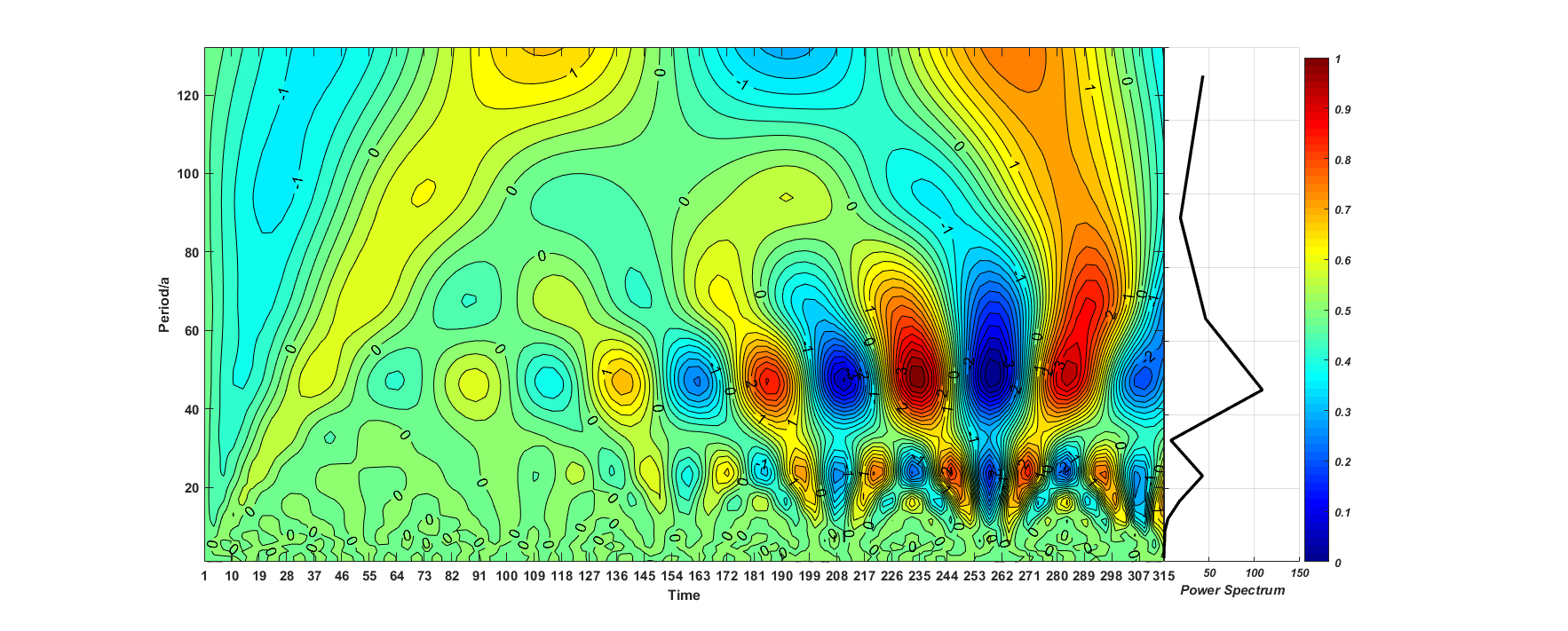}
    \caption{\textbf{Wavelet Analysis of Red Grids Numbers.} Left X-axis: Time in days since simulation. Left Y-axis: Cycle Time Period. Right X-axis: Power Spectrum. Right Y-axis: Intensity}
\end{figure}

Fig.12 shows that the amount of AGH in CA has a period of 46.75 time unit, which meets the stagnation phase of 50 days.
Besides, the only difference between two ODE is the reproduction ability of AGH.

% Require Citation and Explanation
The main factors affecting the reproductive ability of AGH are temperature, humidity, and manual intervention. AGH has high fertility, and a little deviation may cause the amount of the AGH has divergent results. Moreover, AGH is intolerant to frigidity; 
First, AGH almost disappear In the 67th unit which is in winter. However, it has fast outbreak afterwards. Therefore, the Washington Department of Agriculture should pay attention to the number of AGH at next spring. 
Second, 3 unidentified samples were classified as AGH at spring or summer on the west part of Washington. Therefore, we should monitor the potentially spreading at west Washington.

\section{Classification Model}

\subsection{Data Preparation}
% The first figure
\begin{figure}[H]
\centering
\includegraphics[width=16cm]{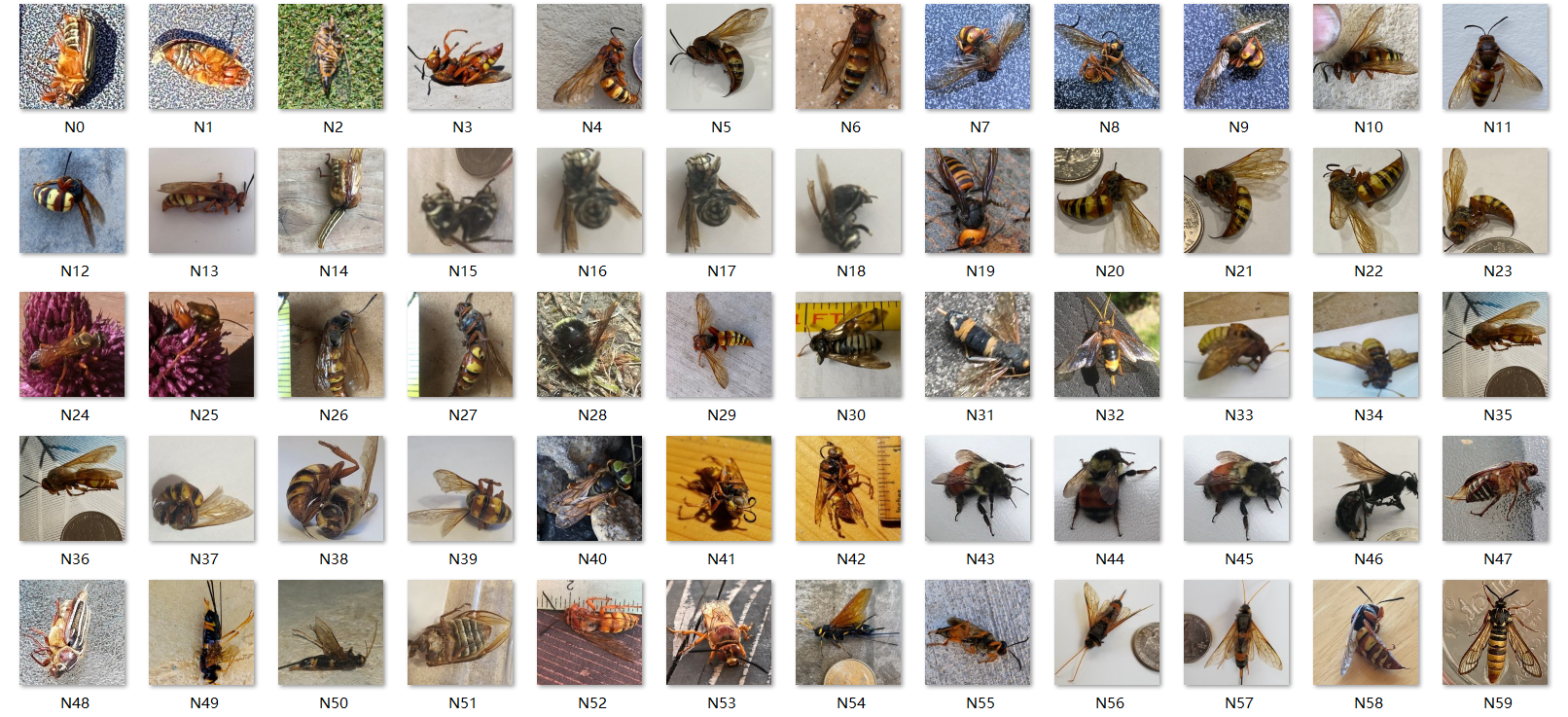}
\caption{\textbf{Batch of Cropped Dataset}}
\end{figure}

We use an AGH dataset from a Github repository \cite{Albert-l-w2021} to make 509 positive samples. We then select 569 negative examples from the given dataset (See Fig.13 for a sample batch). The total amount of pictures for the model training and testing is 1078. We first resize the cropped dataset to the shape of 224*224. Then we add Random Horizontal Flip to flip with a probability of 0.5 horizontally. Next, we standardize the data using ImageNet’s \cite{deng2009imagenet} mean and standard deviation. After that, we split the data into training, validation, and testing set with the ratio of 0.70, 0.15, and 0.15, respectively. We have 757 images for training, 160 for validation, and 161 for testing. Ultimately, we prepare training and testing labels, labeling “1” for positive (AGH) and “0” for negative (non-AGH).

\subsection{Model Construction}
We choose a light-weight model called SqueezeNet \cite{iandola2016squeezenet}. Since our dataset is small (~1000 images), using a light-weight CNN can prevent overfitting and achieve better accuracy. We use pretrained weight from ImageNet. We then change the final classification block output from 1000 to 2 for binary classification. We use Dropout \cite{srivastava2014dropout} with 0.5 probability to prevent overfitting. We also use ReLU \cite{nair2010rectified} activation layers to mitigate vanishing gradient problem.

% Table for SqueezeNet model architecture

\begin{figure}
    \centering
    \includegraphics[width = 17cm]{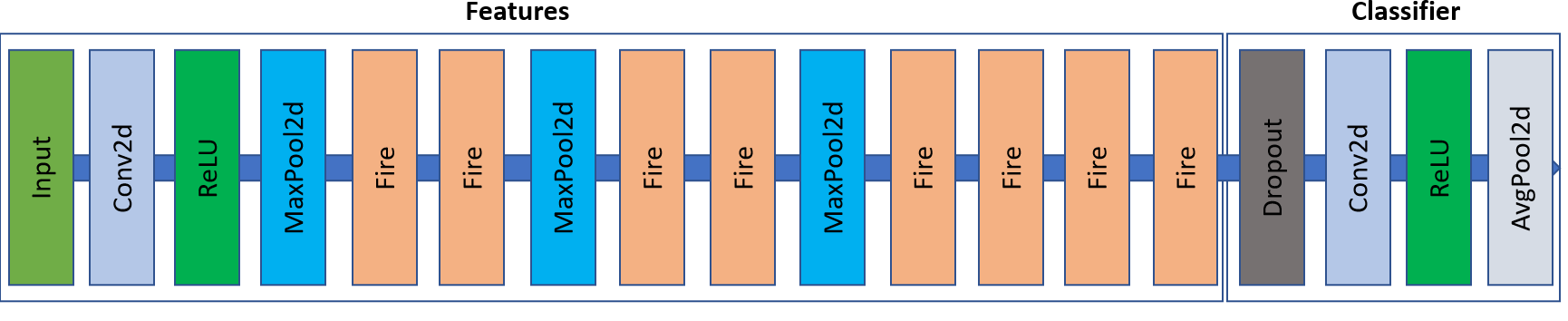}
    \caption{\textbf{Model Architecture of SqueezeNet.} SqueezeNet consists of features layers and classifier layers. We use Conv2d, ReLU, MaxPool2d, Fire, Dropout and AvgPool2d for convolutional layer, ReLU activation layer, maximum pooling layer, Fire Block, Dropout layer and average pooling layer, respectively.}
    \label{fig:my4}
\end{figure}

% Results Detected by SqueezeNet
\begin{figure}[H]
    \centering
    \includegraphics[width = 13cm]{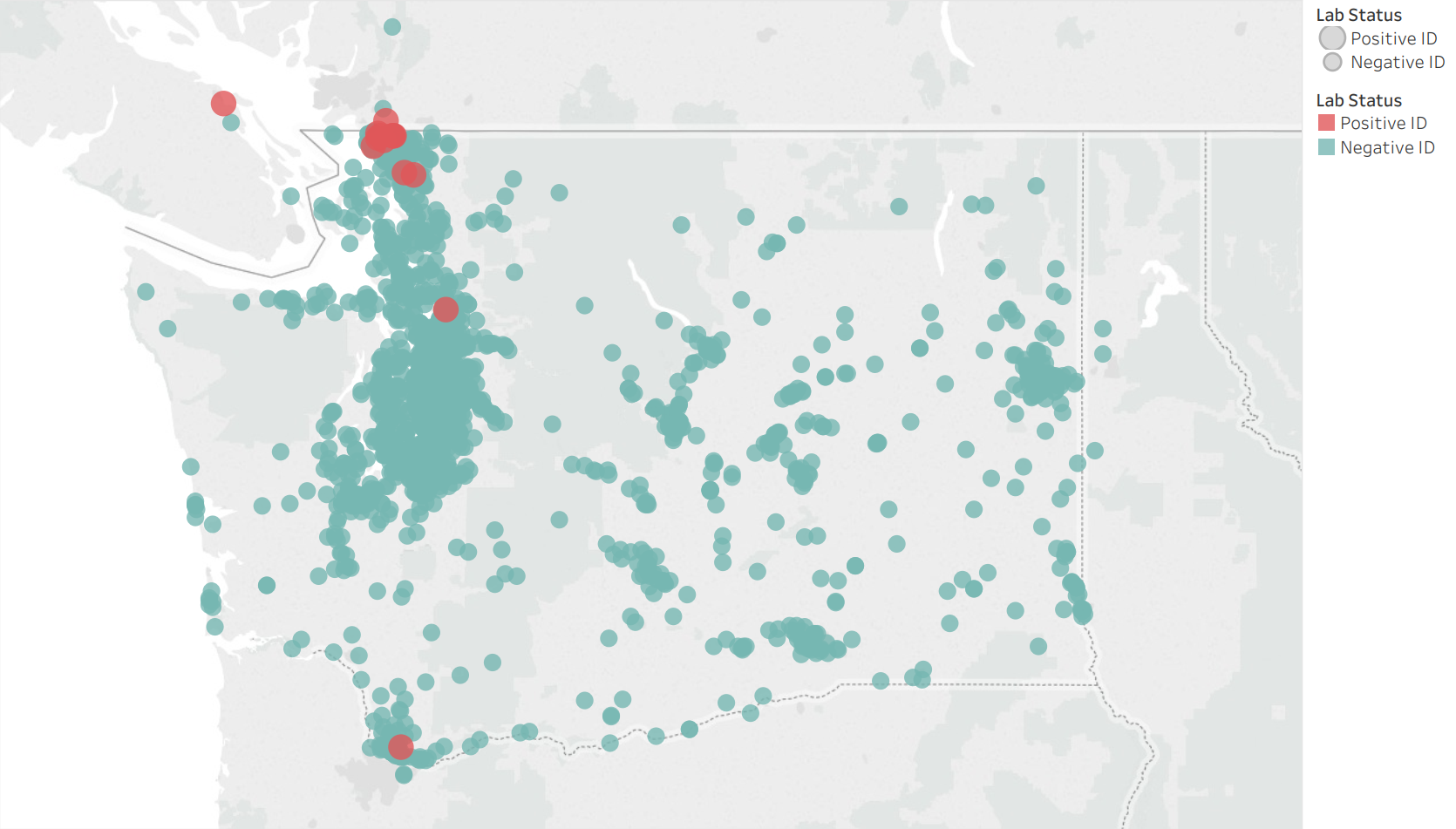}
    \caption{\textbf{Geographical Visualization after SqueezeNet classification.} We use red for positive samples and green for negative samples}
\end{figure}
Fig.15 show most positive samples (AGH) are found at Northwest Washington. Only 1 case is found in Southwest Washington.
We use the Pytorch framework for training our model. We set the epoch as 3 and batch size as 8. We adopt cross-entropy loss as the cost function. We utilize stochastic gradient descent with a learning rate of 1e-3 and momentum of 0.9. Since our training only consists of 3 epochs, there is no need for learning rate decay.
\par

\begin{figure}[H]
\centering
\subfigure[Loss]{
    \includegraphics[width=7cm]{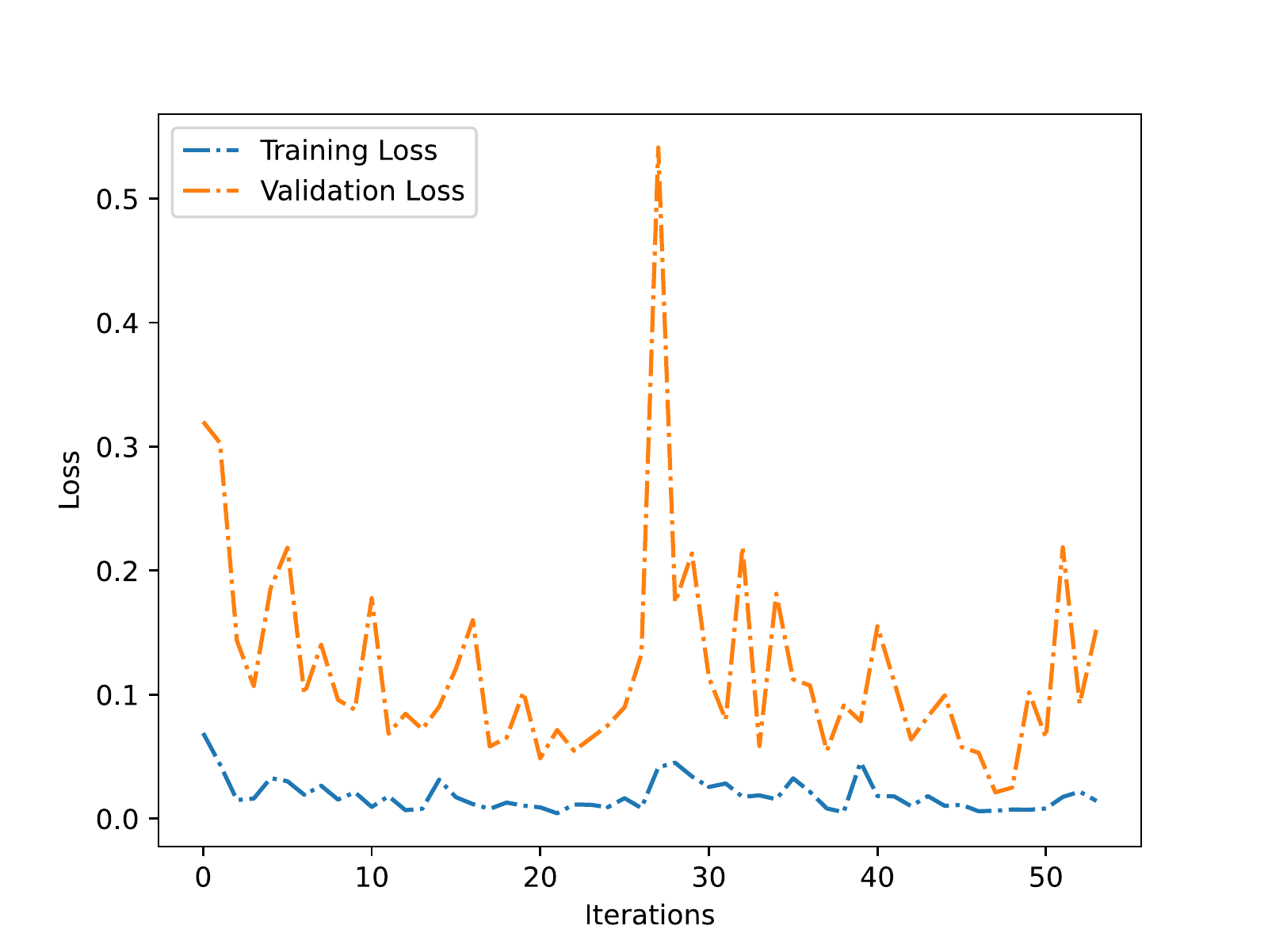}
    }\subfigure[Accuracy]{
    \includegraphics[width=7cm]{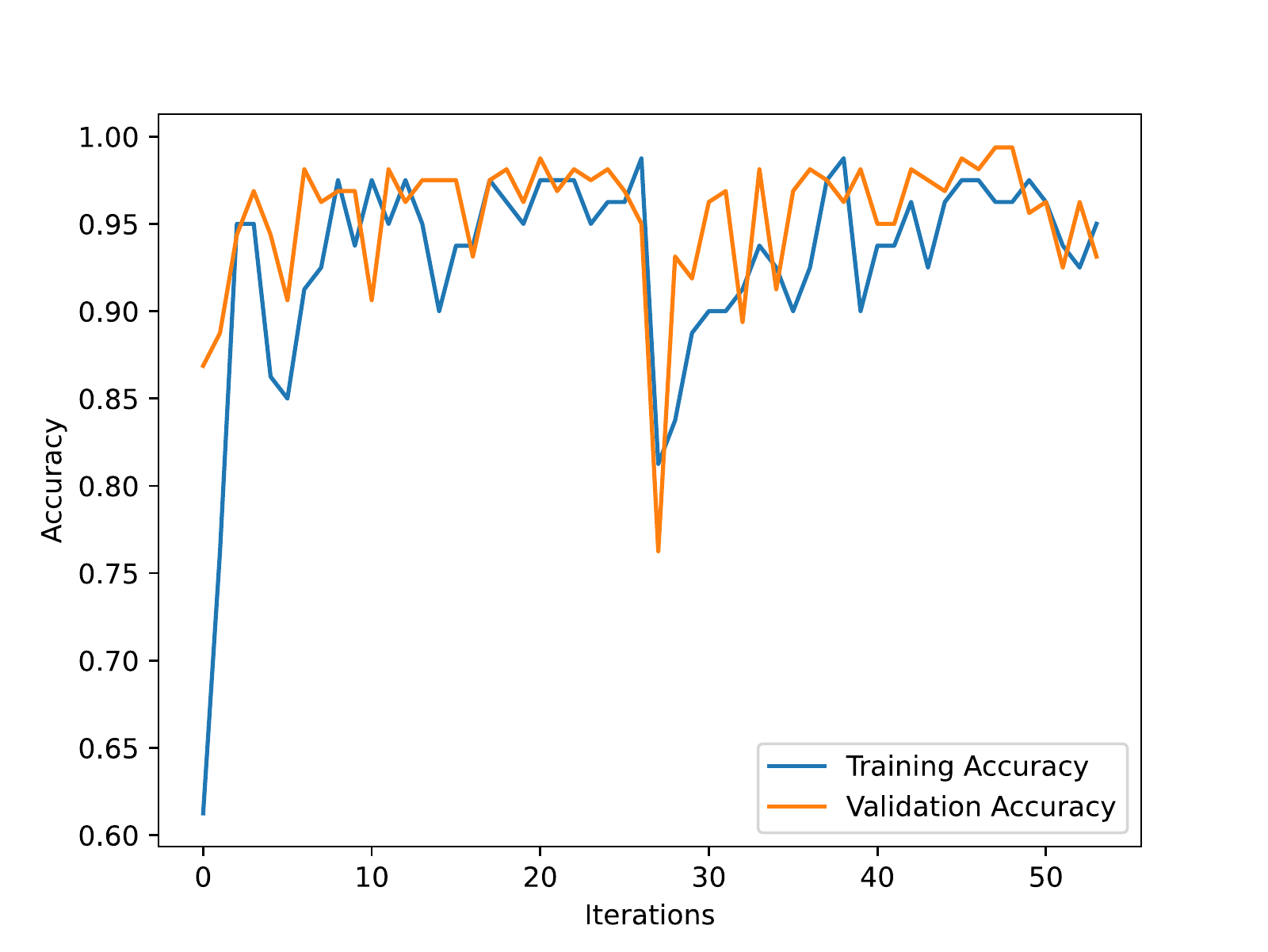}
    }
\caption{\textbf{Loss-Accuracy Plot.} (a): Training and Validation Loss. (b): Training and Validation Accuracy.}
\end{figure}

Fig.16 shows the loss and accuracy during training and validation. Our training loss quickly converges and remains stable between 0 to 0.03. Our validation loss has a spike at the iteration of 30 but has a general trend of declining from 0.3 to 0.1. Except for the sharp decline in the middle, our training and validation accuracy quickly increase and stay between 0.95 and 1.00. Since both accuracies are similar at a high level, there is neither overfitting nor underfitting. We obtain the final testing accuracy of 0.975. 

\subsection{Model Evaluation}
\subsubsection{Evaluation Metrics}
We use a group of evaluation metrics to evaluate our models. Specifically, we include loss, total accuracy, class accuracy, recall, specificity, precision, f1 score, and confusion metric for evaluating our image classification model. 
\par
Confusion matrix displays a table of true positive (TP), False Positive (FP), False Negative (FN) and True Negative (TN). The formula for other evaluation metrics is given as below.
\begin{equation}
\mathrm{Accuracy}=\frac{\mathrm{TP}+\mathrm{TN}}{\mathrm{TP}+\mathrm{TN}+\mathrm{FP}+\mathrm{FN}} 
\end{equation}
\begin{equation}
\operatorname{Recall}=\frac{T P}{T P+F N} 
\end{equation}
\begin{equation}
\text{ Specificity }=\frac{T N}{T N+F P} 
\end{equation}
\begin{equation}
\text { Precision }=\frac{T P}{T P+F P}
\end{equation}
\begin{equation}
F 1 \text { Score }=\frac{2 * \text{Precision} * \text{Recall} }{\text { Precision }+\text { Recall }}
\end{equation}

\subsubsection{Visual Demonstration}

% Activation Layer Result. Positive Samples
\begin{figure}[H]
\centering
\subfigure[Pos1]{
    \includegraphics[width=2.5cm]{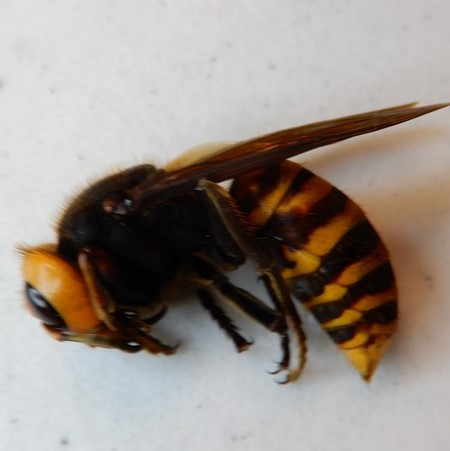}
    }
\subfigure[Pos2]{
    \includegraphics[width=2.5cm]{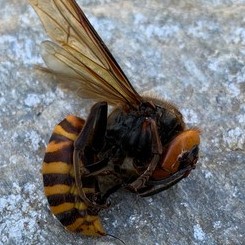}
    }
\subfigure[Pos3]{
    \includegraphics[width=2.5cm]{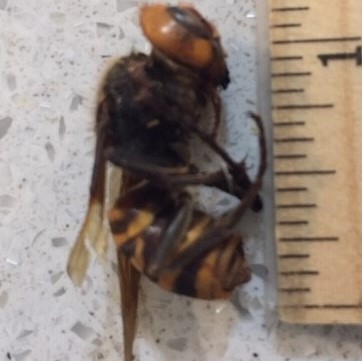}
    }
\subfigure[Pos4]{
    \includegraphics[width=2.5cm]{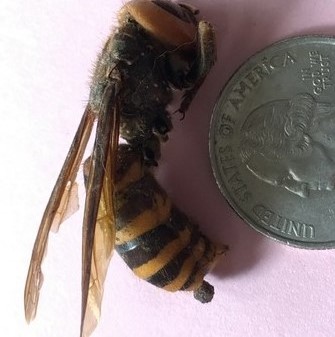}
    }
    \par
\subfigure[Act1]{
    \includegraphics[width=2.5cm]{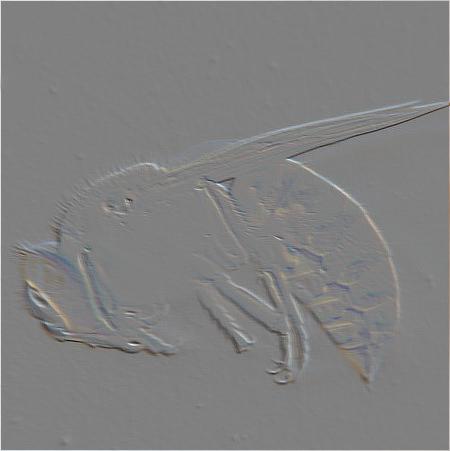}
    }
\subfigure[Act2]{
    \includegraphics[width=2.5cm]{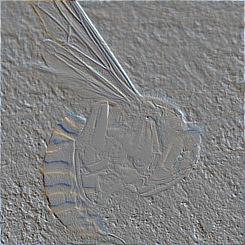}
    }
\subfigure[Act3]{
    \includegraphics[width=2.5cm]{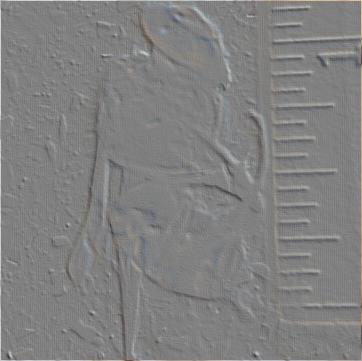}
    }
\subfigure[Act4]{
    \includegraphics[width=2.5cm]{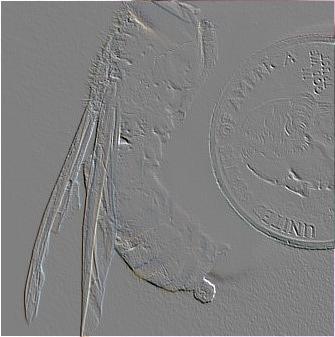}
    }
\caption{\textbf{Activation Layers Result with samples classified as positive (AGH).} The upper row is the input image. The lower row is the output from the first convolution layer. We use "Pos" for positive samples and "Act“ for activation layers, respectively.}
\end{figure}

% Activation Layer Result. Negative Samples

\begin{figure}[H]
\centering

\subfigure[Neg1]{
    \includegraphics[width=2.5cm]{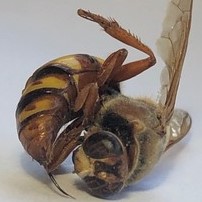}
    }
\subfigure[Neg2]{
    \includegraphics[width=2.5cm]{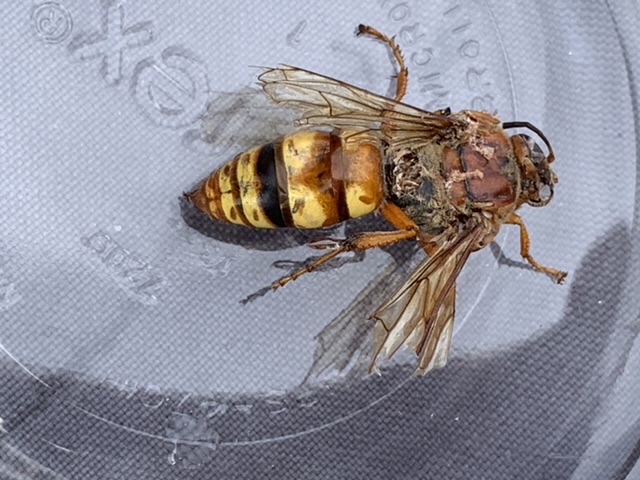}
    }
\subfigure[Neg3]{
    \includegraphics[width=2.5cm]{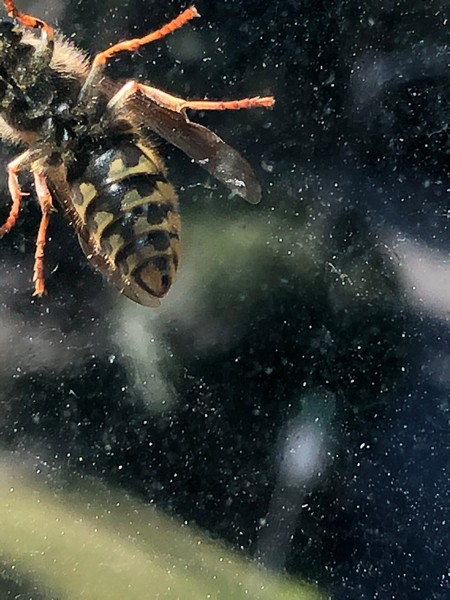}
    }
\subfigure[Neg4]{
    \includegraphics[width=2.5cm]{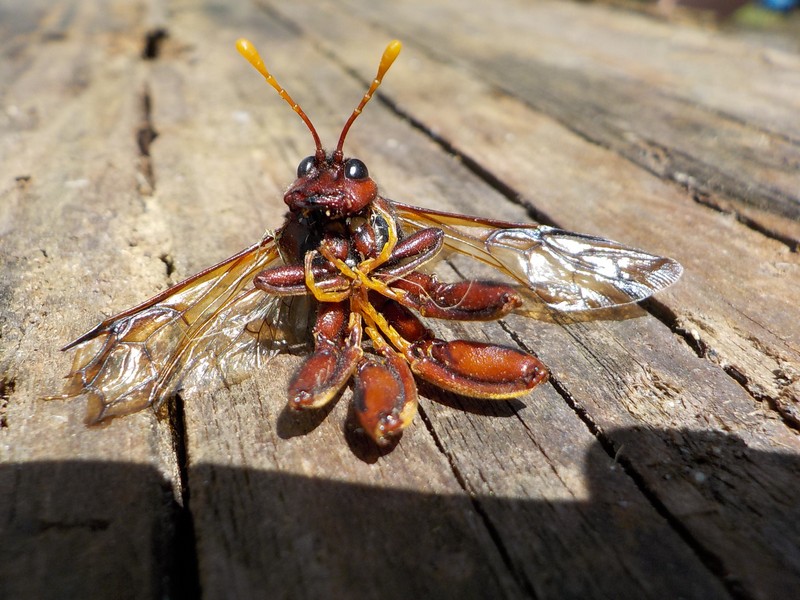}
    }
    \par
\subfigure[Act1]{
    \includegraphics[width=2.5cm]{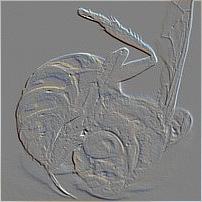}
    }
\subfigure[Act2]{
    \includegraphics[width=2.5cm]{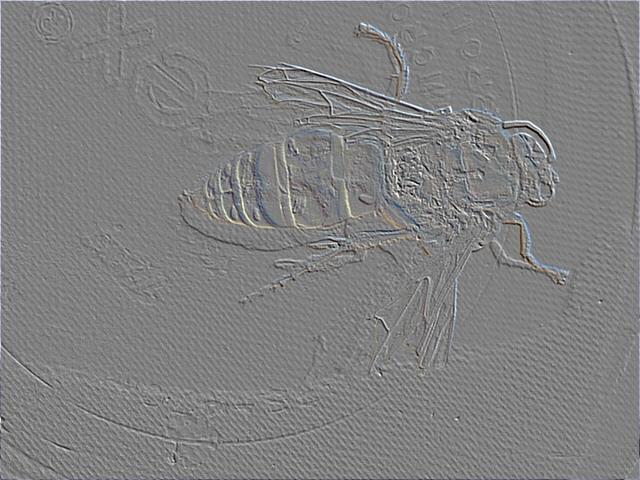}
    }
\subfigure[Act3]{
    \includegraphics[width=2.5cm]{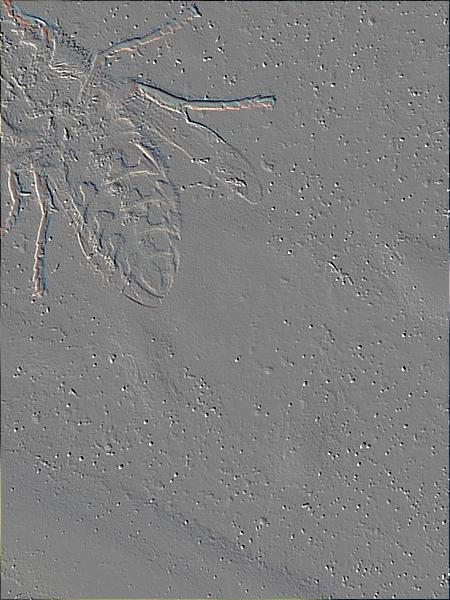}
    }
\subfigure[Act4]{
    \includegraphics[width=2.5cm]{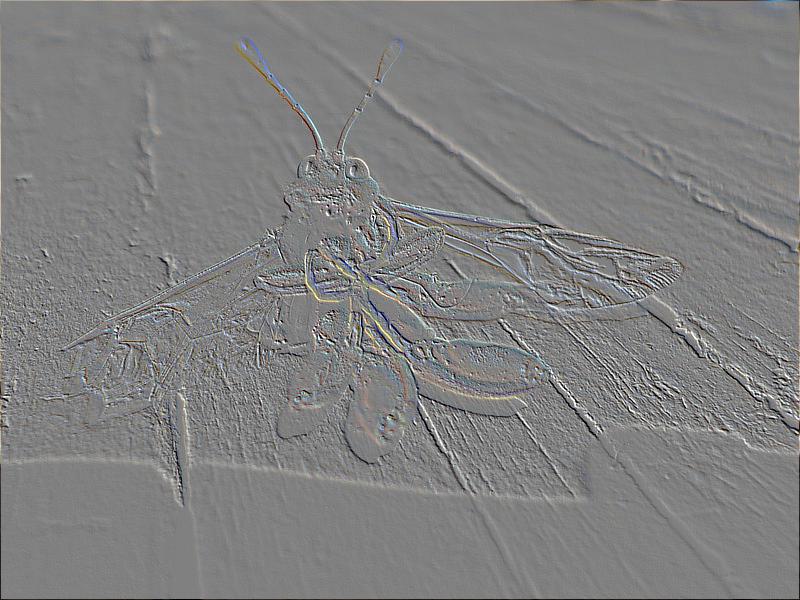}
    }
\caption{\textbf{Activation Layers Result with samples classified as Negative (non-AGH).} The upper row is the input image. The lower row is the output from the first convolution layer. We use "Neg" for positive samples and "Act“ for activation layers, respectively.}
\end{figure}

We visualize the first convolutional layer result in Fig.17 and Fig.18. Our rows display the positive sample, positive sample activation result, negative sample, and negative sample activation result, respectively. As shown in the figure, our neural network can filter background interference and separate the given insects' pattern. Besides, our CNN displays more evident edges for positive samples. That means our model can efficiently capture the distinct feature of the AGH.

\subsubsection{Quantitative Comparison}
In this part, we quantitatively analyze our model classification performance. We first display the confusion matrix. According to Fig.19, we have 82 true positives, 75 true negatives, 3 false positives, and 1 false negative. We also note from Table 3 that all evaluation metrics have a score greater than 0.96. That means our model has great classification accuracy. Besides, our model has better predictive accuracy at positive samples than negative samples. That is reasonable because the negative examples contain numbers of insects that look similar to AGH (e.g., European Hornet). Table 5 shows the image quality. Positive samples win all evaluation metrics except image contrast. That means image contrast is a misleading metric for classifying AGH. 

% Table for quantitative result
\begin{table}[h]
\centering{
\begin{tabular}{lllllll}
\toprule[1.5pt]
 & \textbf{Accuracy} & \textbf{Class Accuracy} & \textbf{Recall} & \textbf{Specificity} & \textbf{Precision} & \textbf{F1 Score}
\\
\hline
Value   & 0.975    & 0.965/0.987    & 0.988  & 0.962       & 0.965     & 0.976 
\\
\toprule[1.5pt]
\end{tabular}
}
\caption{\textbf{Quantitative result of our image classification model.} Class Accuracy has negative (not AGH) on the left and positive (AGH) on the right.}
\end{table}

% the plot for confusion matrix
\begin{figure}[H]
\centering
\includegraphics[width=6cm]{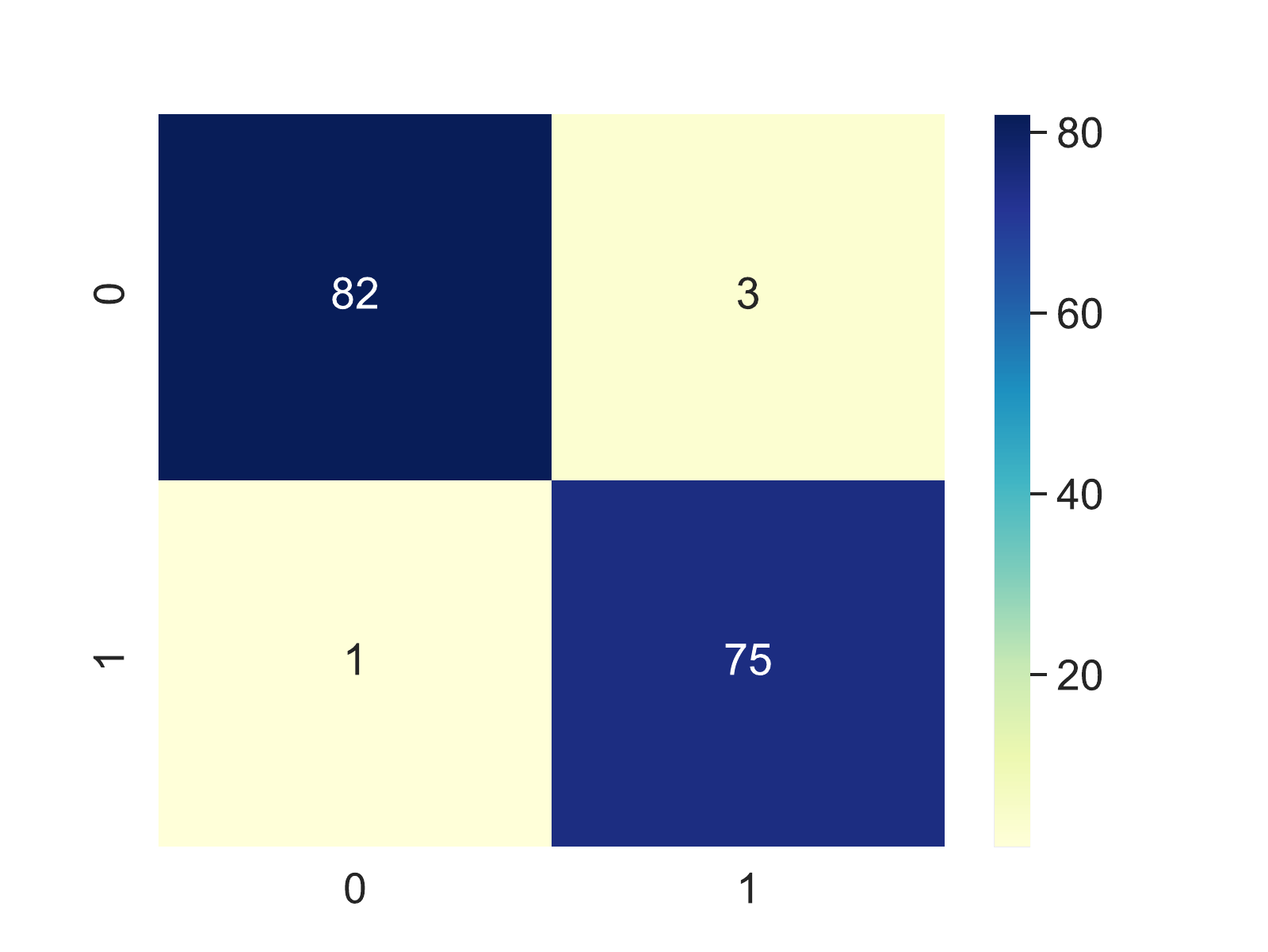}
\caption{\textbf{Confusion Matrix of our image classification model.} Left: Actual Label. Bottom: Predicted Label. from upperleft to lowerright cell: True Positive, False Positive, False Negative and True Negative, }
\end{figure}

\subsection{Image Evaluation}
\subsubsection{Image Quality Metrics}
We use a set of non-reference image quality metrics to understand why people would misclassify other insects as Asian hornets. That group of image quality metrics would also help investigators screen out a more likely picture to contain AGH. Our image quality metrics include Naturalness Image Quality Evaluator (NIQE) \cite{mittal2012making} , image gradient, entropy, and contrast. We also use TOPSIS to summarize the indicators above. NIQE is a statistical measure derives from natural, undistorted images. Image Gradient is a vectorized change in image intensity. Image Entropy is a statistical measure of randomness for assessing the texture of an image. Image Contrast is the difference in luminance that makes a picture visually distinguishable. TOPSIS is a multi-criteria decision analysis method that uses weighted and normalized geometric distance for assessing image quality.
We use TOPSIS to give a comprehensive evaluation of image quality.
\\
NIQE:
\begin{equation}
    \begin{aligned}
D\left(\nu_{1}, \nu_{2},\right.&\left.\Sigma_{1}, \Sigma_{2}\right) \\
&=\sqrt{\left(\left(\nu_{1}-\nu_{2}\right)^{T}\left(\frac{\Sigma_{1}+\Sigma_{2}}{2}\right)^{-1}\left(\nu_{1}-\nu_{2}\right)\right)}
\end{aligned}
\end{equation}
Image Gradient:
\begin{equation}
    \nabla f=\left[\begin{array}{l}
g_{x} \\
g_{y}
\end{array}\right]=\left[\begin{array}{l}
\frac{\partial f}{\partial x} \\
\frac{\partial f}{\partial y}
\end{array}\right]
\end{equation}
Entropy:
\begin{equation}
    S=k_{b} \ln \Omega
\end{equation}
Contrast:
\begin{equation}
    \sum_{i=0}^{N-1} \sum_{j=0}^{M-1}\left(I_{i j}-\bar{I}\right)^{2}
\end{equation}
\\
We acquire the positive and negative ideal points of each indicator in TOPSIS:
\begin{equation}
    d_{i}^{*}=\sqrt{\sum_{j=1}^{4}\left(z_{i j}-z_{j}^{*}\right)^{2}}, d_{i}^{-}=\sqrt{\sum_{j=1}^{4}\left(z_{i j}-z_{j}^{-}\right)^{2}}
\end{equation}
where Z is the weighted evaluation matrix.

\begin{table}[]
\centering{
\begin{tabular}{lllll|l}
\toprule[1.5pt]
         & NIQE & Gradient & Entropy & Contrast & TOPSIS \\ 
         \hline
Negative & 6.98  &  0.057 &  7.385  & \textbf{3092}  &  0.1912      \\ 
\hline
Positive & \textbf{10.10}  &  \textbf{0.075} &  \textbf{7.525}  & 2838  &  \textbf{0.2655}      \\ \toprule[1.5pt]
\end{tabular}
}
\caption{\textbf{Image Quality Comparison for positive and negative samples.} We use bold to indicate higher values.}
\end{table}

\subsubsection{Model Implementation}
We use four non-reference image quality metrics values as features and the other one as the label. We then split the training and testing data with a train/test ratio of 2:1 and used random forest classifiers to classify positive and negative samples. 
\par
Random Forest classifier is an ensembled supervised machine learning algorithm. It is a meta estimator that leverages numbers of decision trees and takes the average of those predictors to improve forecasting accuracy and reduce predictability variability. 

\begin{table}[h]
\centering{
\begin{tabular}{lllll}
\toprule[1.5pt]
& \textbf{n\_estimators} & \textbf{max\_depth} & \textbf{min\_samples\_split} & \textbf{min\_samples\_leaf} \\
\hline
Value           & 644           & 110        & 10                  & 2  \\
\toprule[1.5pt]
\end{tabular}
}
\caption{\textbf{Best Hyperparameters of the random forest classifier.} n\_estimators: numbers of decision tree in random forest. max\_depth: maximum depth of the tree.min\_samples\_split: minimum samples for splitting internal nodes. min\_samples\_leaf: minimum samples for a leaf node.}
\end{table}

We initially set the min sample split as two and the numbers of estimators as 150. We then select four hyperparameters for random grid search cross-validation. Specifically, We run 3-fold cross-validation for 100 iterations and find the hyperparameter group that minimizes the loss function. The best hyperparameters summarize in the Table 5. Our Testing Accuracy from the optimized hyperparameters is approximately 0.686.

\subsubsection{Feature Importance \& Error Analysis}
% Feature Importance score
\begin{figure}[H]
\centering
\subfigure[]{
\includegraphics[width=8.2cm]{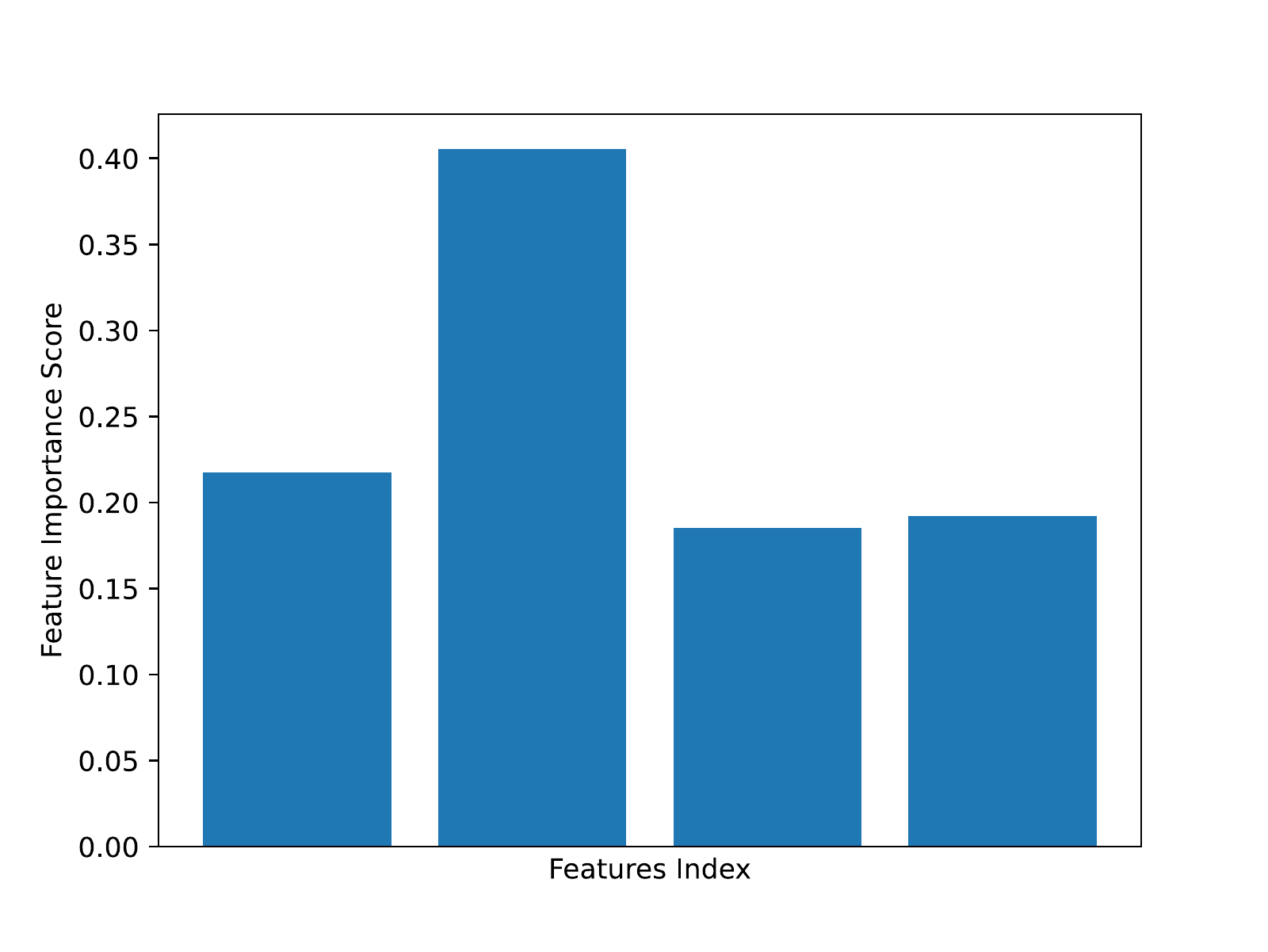}
}
\subfigure[]{
\includegraphics[width=8.2cm]{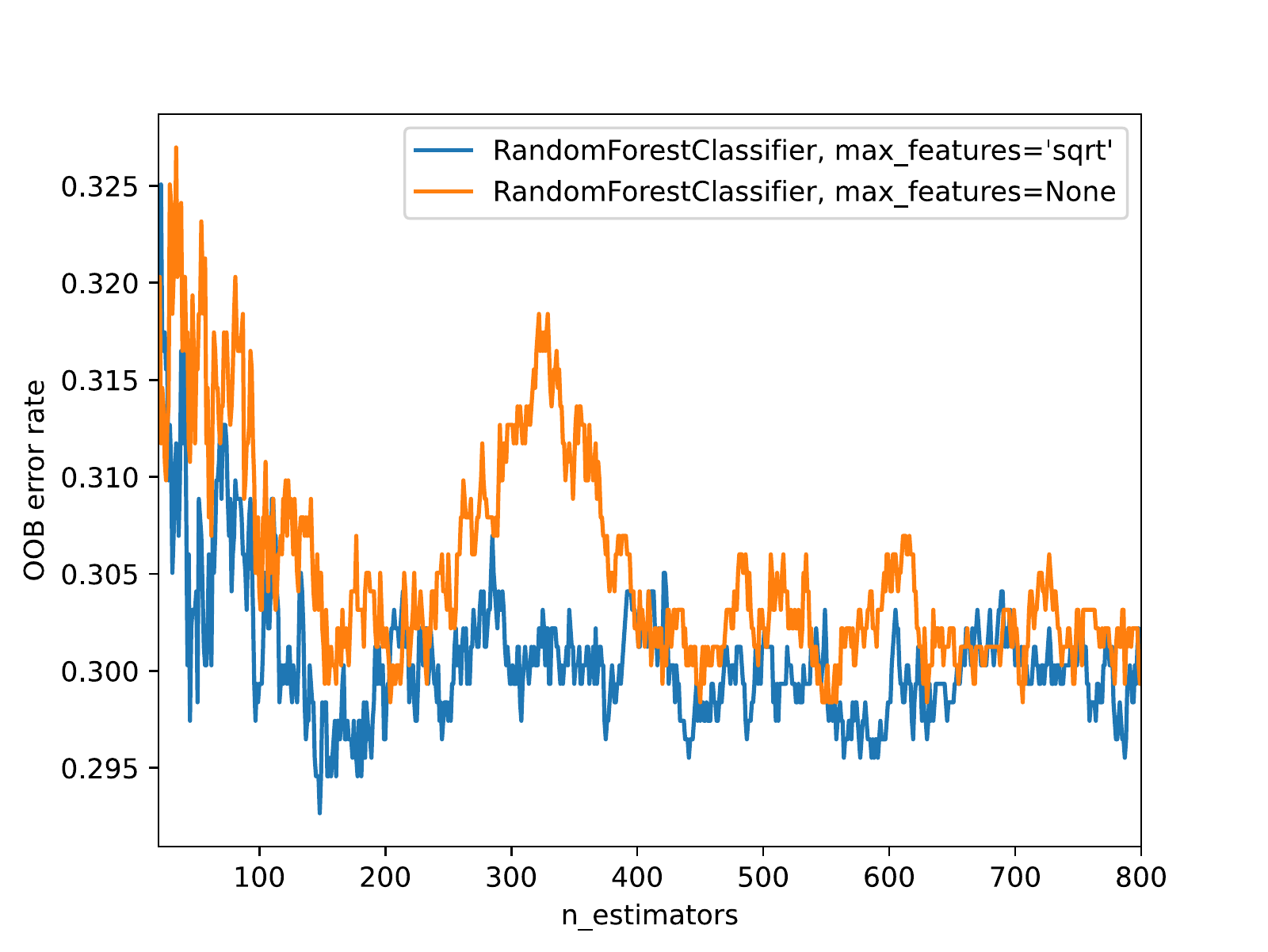}
} 
\caption{(a): \textbf{Feature Importance Score}. X-axis: feature index. From left to right, NIQE, gradient, entropy and contrast. Y-axis: Feature Importance Score, from 0 to 1. (b): \textbf{Error Analysis}. X-axis: number of estimators. Y-axis: out-of-bag (OOB) error rate}
\end{figure}

We visualize the feature importance in the Fig.20(a). The feature importance score describes how significant a specific feature is for image classification. We observe from the plot that the image gradient has dominant importance (around 0.40) for classification, whereas other factors contribute similarly. 

We also examine the source of classification errors in Fig.20(b). In Fig.20 (b), we note different types of maximum features: the number of features for calculating the best split. When max features are 'sqrt' (the blue line), the classifier will consider maximum features as the total features' square root. When max features are 'None' (the yellow line), it equates maximum features to total features.

For both max features types, we have decreasing OOB error rate with increasing numbers of features. However, we note that when numbers of estimators grow greater than 600, the OOB error rate remains stable. Therefore, it is efficient to stop at around 600.

\section{Sensitivity Analysis}
\subsection{Expansion Model}
\begin{figure}[H]
\centering
\subfigure[0.475<a<0.525]{
\includegraphics[width=8cm]{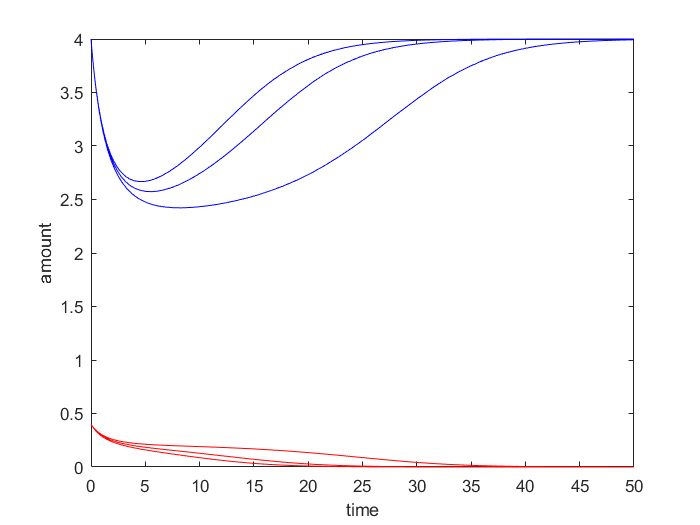}
}
\quad
\subfigure[0.38<c<0.42]{
\includegraphics[width=8cm]{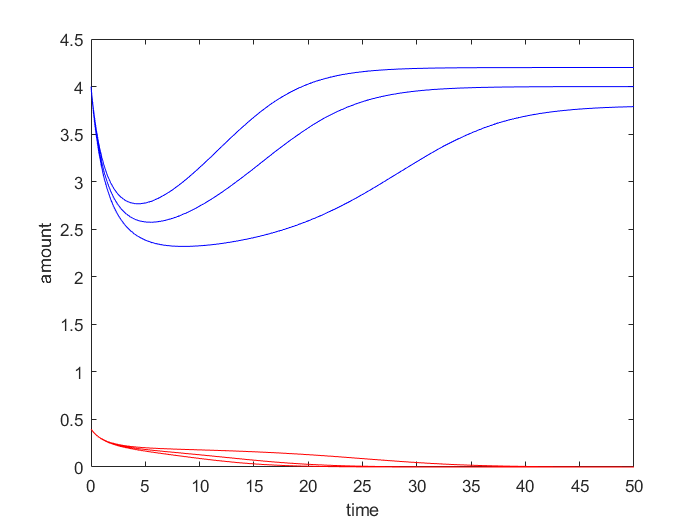}
}
\quad
\subfigure[0.19<b<0.21]{
\includegraphics[width=8cm]{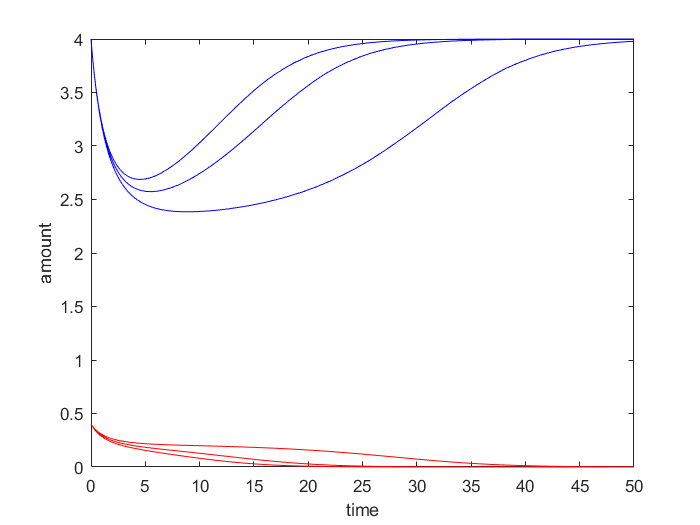}
}
\quad
\subfigure[0.76<e<0.84]{
\includegraphics[width=8cm]{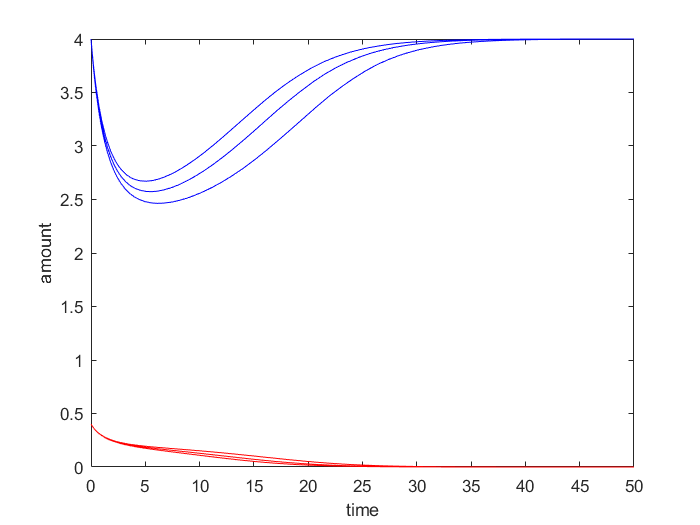}
}
\caption{\textbf{Expansion Model Response to Input Variation}}
\end{figure}

Fig.21 shows how our expansion model responds to input variation. For AGH, Our model has reasonable stability under input variation. For other insects, The internal competition coefficient of AGH results in the most significant fluctuation, whereas the internal competition coefficient of local species leads to the least fluctuation. 

\subsection{Classification Model}
% Blurred Images
\begin{figure}[H]
\centering
\subfigure[Ground Truth]{
    \includegraphics[width=3cm]{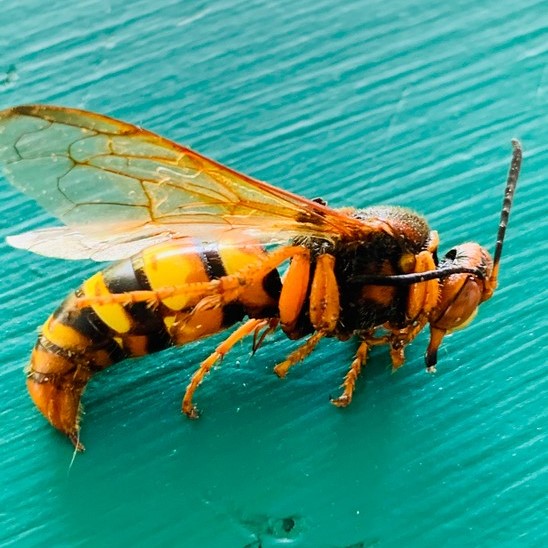}
    }
\subfigure[Averaging Blur]{
    \includegraphics[width=3cm]{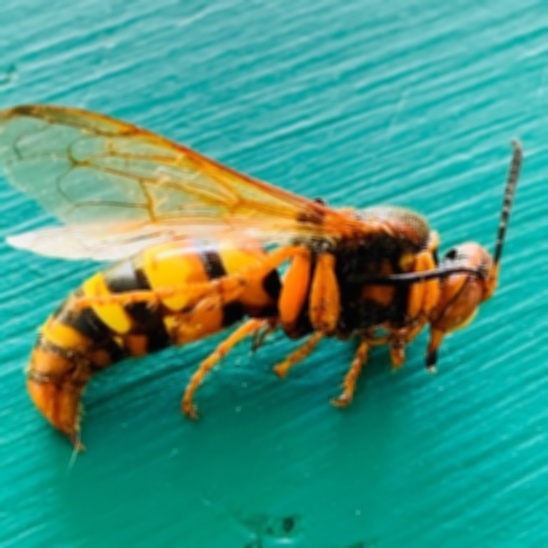}
    }
\subfigure[Gaussian Blur]{
    \includegraphics[width=3cm]{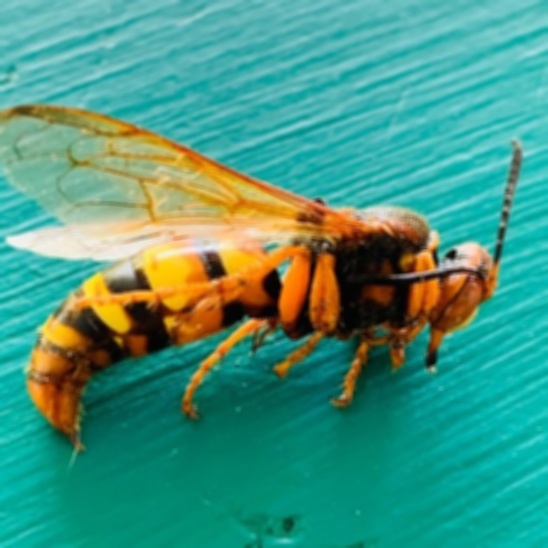}
    }
\subfigure[Median Blur]{
    \includegraphics[width=3cm]{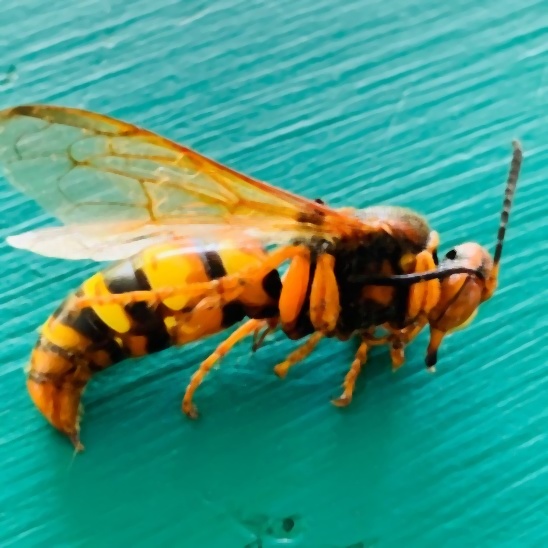}
    }
\caption{\textbf{Clean and Blurred Images. All blur has kernel size of 5.}}
\end{figure}

In this subsection, we conduct sensitivity analysis to the classification model. We first apply disruptive factors, including averaging, Gaussian Blur, and Median Blur, to the groudtruth. The resulting images are in Fig.22.

% Confusion matrix comparison Plot
\begin{figure}[H]
\centering
\subfigure[Average Blur]{
    \includegraphics[width=4.5cm]{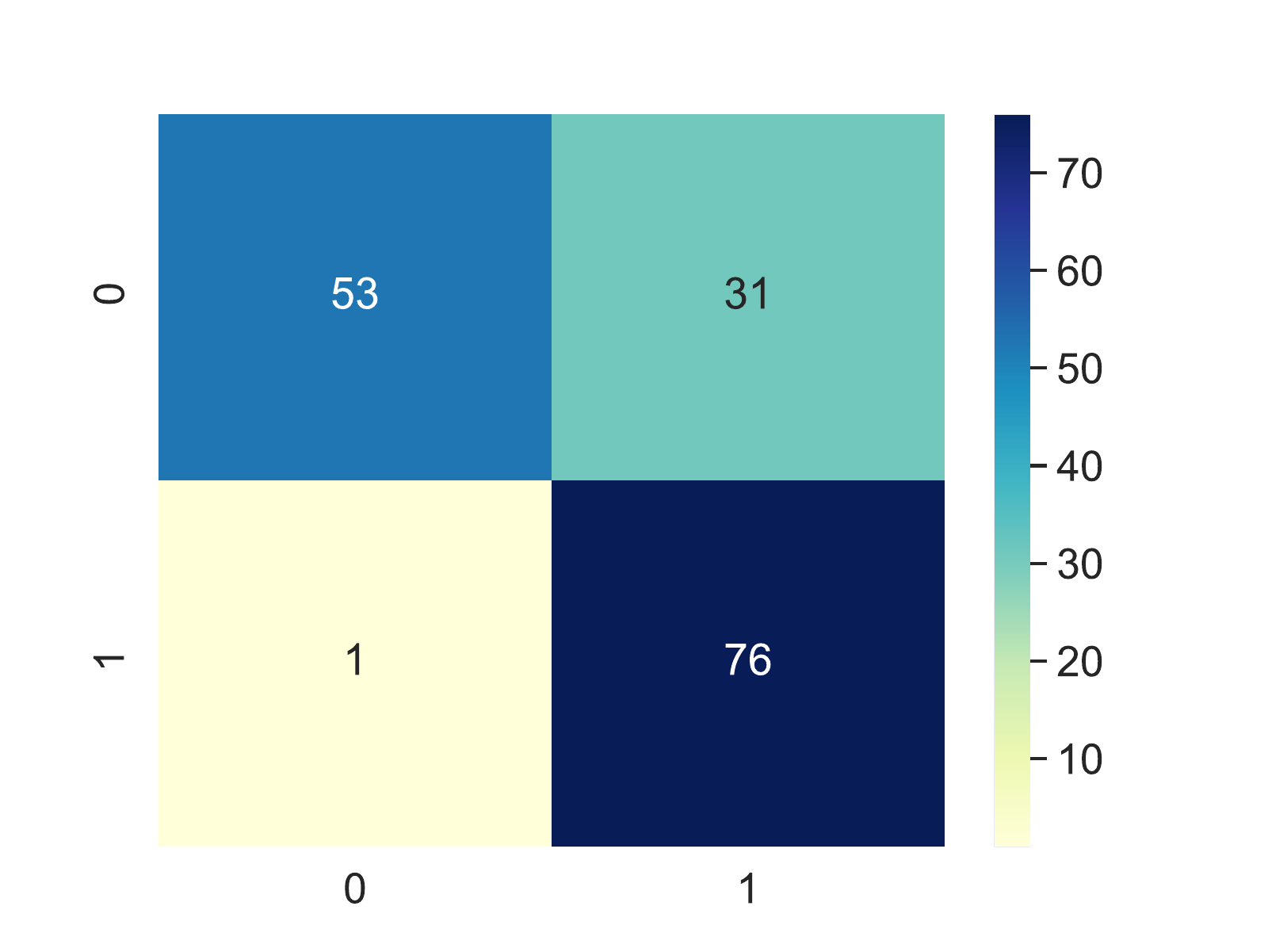}
    }
\subfigure[Gaussian Blur]{
    \includegraphics[width=4.5cm]{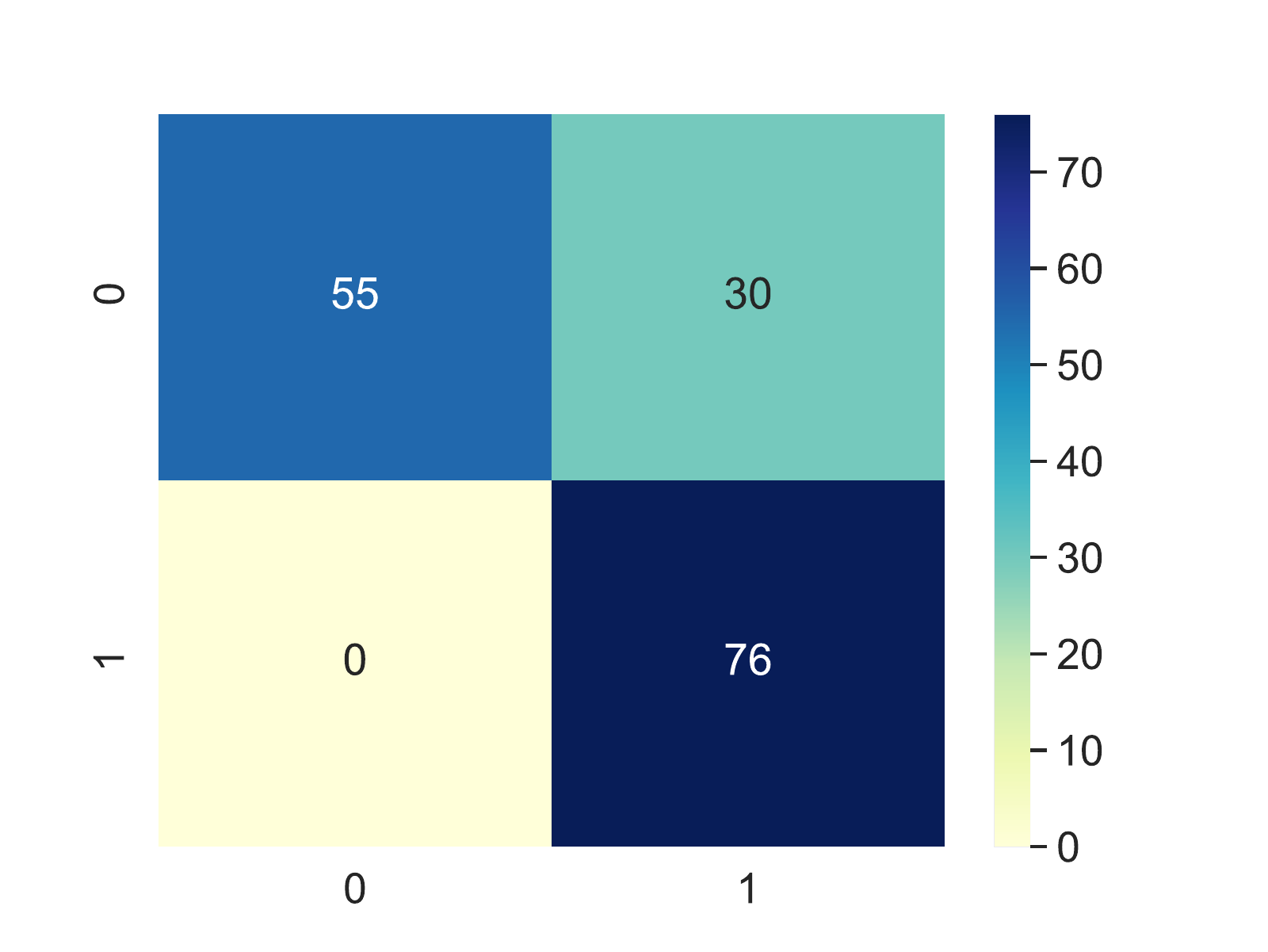}
    }
\subfigure[Median Blur]{
    \includegraphics[width=4.5cm]{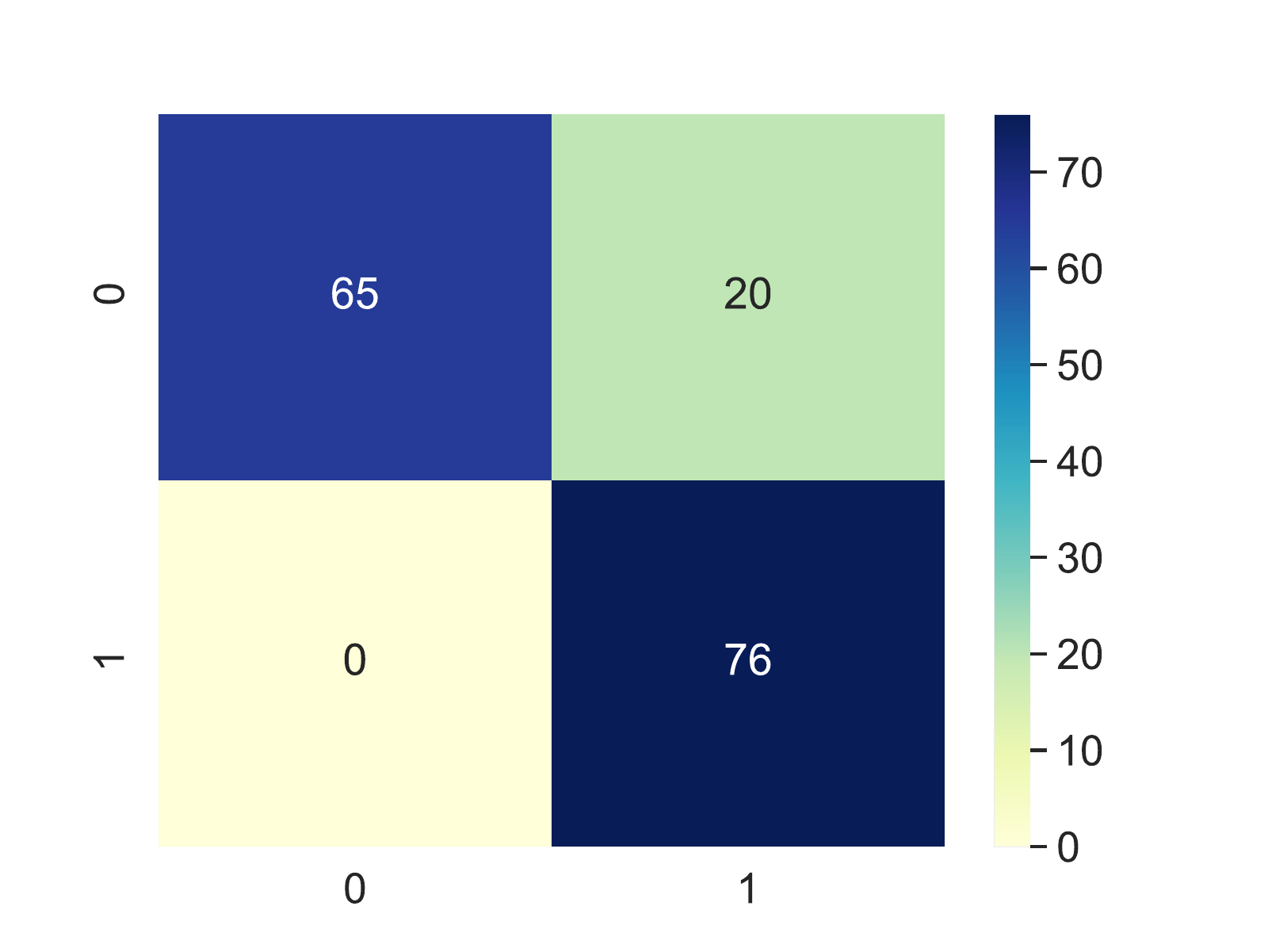}
    }

\caption{\textbf{Confusion matrix under different blur conditions}}
\end{figure}

In Fig.23, We display the confusion matrix under different blur conditions. Under all blur conditions, our model has 0 or only 1 false negatives. Meanwhile, it has some false positives, but its number is significantly less than true positives and true negatives. That means our model can recognize Asia Wasps under disturbance.

% The Radar Plot
\begin{figure}[H]
\centering
\includegraphics[width=10cm]{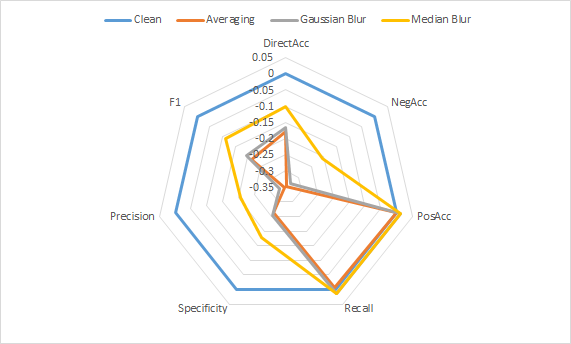}
\caption{\textbf{Radar Plot under different blur conditions}}
\end{figure}

We then build a radar plot in Fig.24 to assess model performance under disturbance. All blurring has no significant influence on positive accuracy and recall. However, Averaging Blur and Gaussian Blur will slightly decrease direct accuracy, specificity, and f1 score and significantly reduce negative accuracy and precision. Meanwhile, median blur results in less degradation. 

\section{Model Reflection}

\subsection{Strengths}
\begin{itemize}
% Pros for data preprocessing

% Pros for Differential Equation 

    \item 
    Consistent classification and predictive ability.
    \item
    Quick and Convenient Model Update from additional data.
    \item
    High Simulation Accuracy with seasonal periodic change, Brownian motion and potential propagation coefficient 

% Pros for Cell Automation Machine
\item Efficient Analysis of complex images and videos
% Pros for convolutional neural network 
\item Great classification performance and robustness
% Pros for random forest classifier & non-reference metrics
\item Novel binary classification from pure image quality.
\end{itemize}
\subsection{Weaknesses}
\begin{itemize}
    \item 
    Idealize inter-species competition and override human intervention factors.
    \item 
    Possibly inconsistent potential propagation coefficient update in cellular automation 

% Cons for data preprocessing

% Cons for Differential Equation 
\item Over-Simplified factors others than dual species competition between AGH and Bees.

% Cons for Cell Automation Machine

% Cons for convolutional neural network 
\item Sub-optimal interpretability at intermediate layer's feature processing
% Cons for random forest classifier & non-reference metrics
\item Subjective Image Indicator for binary classification. 
\end{itemize}

\clearpage
\section{Memorandum}
\textbf{To:} The Washington State Department of Agriculture\par
\textbf{From:} Team \#2116473 of 2021 MCM\par
\textbf{Date:} Feb 9, 2021\par
\textbf{Subject:} Recommendations and Solutions to Potential Bioinvasion of Asian Giant Hornet\par
~\\~\\
Honorable Washington State Department of Agriculture,
~\\\par
Recently, Asian giant hornet was discovered around Washington State. According to research, this species of hornet could be extremely dangerous and a serious threat that will probably lead to a biological invasion in a short time. To predict the pest's spread and control it in the future, we make a serial of analysis upon current data.
~\\\par
The report has analyzed the Asian giant hornet using statistics, mathematics, and deep learning tools. First, we visualize the geographical distribution of insects, investigate insect populations of different dates and examine the their expansion period. Second, we calculate Asian giant hornet numbers at different conditions and simulate its geographical spread. Third, we build an advanced classification model and discuss the reason for widespread mistaken reporting.  Besides, we demonstrate the robustness of our models under various input variations. Finally, we recommend measures for successfully addressing Asian giant hornet.
~\\\par
There are insightful findings from our analysis.
\begin{itemize}
    \item 
    Most Asian giant hornet is reported at west part of Washington. We use a deep learning model with 0.975 testing accuracy for verification. However, we only found 14 Asian Giant Hornet among 4400+ reports.
    \item 
    Local insects have a monthly active cycle of about 7.71 days. Asian giant hornet breed at a rapid rate every Spring.
    \item
    Asian giant hornet's explosion rely on high potential propagation coeffcient, according to our geographical spread analysis. However, AGH are inactive all year along and can hardly reach that condition.
    \item 
    Asian giant hornet's fragile at early explosion. Timely human interference (about 30 days earlier than spreading) could efficiently control the AGH population and protect other insects.
    \item 
    Light-weight deep learning models is far more accurate (>0.97) for classifying Asian giant hornet than human observations (<0.01).
    \item
    It is most efficient to update our models (with pretrained weight saved) after every winter. 
    \item 
    Reported image with high intensity change is a good metric for subjectively classifying AGH, whereas images with high luminary is misleading.

\end{itemize}

We finally provide recommendations based on our analysis result.
\begin{itemize}
    \item 
    Clear Asian giant hornet hives at early spring. You may use fire fighting drones to accelerate that process. 
    \item 
    Timely Classify and Report Asian giant hornet using light-weight convolutional neural network. Once you detect the signs of an outbreak, you must take action within 30 days.
    \item 
    Ignore reports that has low image intensity change or has noisy backgrounds. They are almost certainly fake Asian giant hornet images. Be careful of misguiding image with high contrast.
    \item 
    Update classification and explosion models at every winter. At that time, you should have collected data till the end of a year.
    \item 
    There is no need for massive panic about Asian giant hornet. Even among large numbers eyewitnesses reports over a long period of time, there are few AGH.
    \item
    The public should be informed of AGH to reduce unnecessary review time. Meanwhile, you could prioritize images with high TOPSIS score.
\end{itemize}
 \par We suggest exploiting updating our model in the following ways: ~\\\par
 First, you can use the non-reference image quality metrics and Topsis score to determine the order of image audit. Then you can use SqueezeNet to identify if a image contains an Asian giant hornet. For images that are difficult to discriminate, you can use manual screening.~\\\par
After screening the images, you should collect the data on the adaptation of Asian Giant Hornets in Washington to update the differential equations for simulation. You should also pay attention to the number of Asian giant hornet, at early Spring. Finally, the government should take prompt action when there is a evidence of outbreak.
~\\\par
We hope our analysis and advice will help to control spreading of Asian giant hornets and generating better plans for future issues. You are welcome to contact us at any time.\par
\setlength{\parindent}{0cm}
~\\~\\
Yours Sincerely,\par
\setlength{\parindent}{0cm}
Team \#2116473 of 2021 MCM

\clearpage
\bibliographystyle{unsrt}
\bibliography{my}

%\hspace{2em}
%\begin{appendices}

%\end{appendices}

\end{document}